\documentclass[lettersize,journal]{IEEEtran}
\usepackage{amsmath,amsfonts}
\usepackage{algorithmic}
\usepackage{algorithm}
\usepackage{array}
\usepackage{subcaption}
\usepackage{textcomp}
\usepackage{stfloats}
\usepackage{url}
\usepackage{verbatim}
\usepackage{graphicx}
\usepackage{cite}
\usepackage{subcaption} 
\usepackage{booktabs}
\usepackage{multirow}
\usepackage{subfig}
\usepackage{siunitx}

\usepackage{amsmath}
\begin{document}

\title{Mitigating Sample-Level Imbalance via Probabilistic Separation for Adaptive Multimodal Fusion}

\author{Zhiwen Yu,~\IEEEmembership{Senior Member,~IEEE,} Zhaocheng Liu, Xiaoqing Liu,~\IEEEmembership{Student Member,~IEEE,} Huanqiang~Zeng,~\IEEEmembership{Senior Member,~IEEE,} C. L. Philip Chen,~\IEEEmembership{Life Fellow,~IEEE}

\thanks{This work was supported in part by the National Natural Science Foundation of China No. 6257070404 and 92467109, in part by National Key R\&D Program of China 2023YFA1011601, in part by the Major Key Project of PCL (Grant No. PCL2025A11 and No. PCL2025A13), in part by Guangzhou Science and Technology Plan Project 2024A04J3749, in part by National Natural Science Foundation of China 62106224, U21A20478, and in part by Guangdong Basic and Applied Basic Research Foundation under Grant 2024A1515140137. (Corresponding author: Zhiwen Yu)}
\thanks{Zhiwen Yu, Zhaocheng Liu and C. L. Philip Che are with the School of Computer Science and Engineering in South China University of Technology, Guangzhou 510006, China (e-mail: zhwyu@scut.edu.cn, cszhch60216@mail.scut.edu.cn, Philip.Chen@ieee.org).}
\thanks{Xiaoqing Liu is with the School of Future Technology in South China University of Technology, Guangzhou 510641, China, and also with Pengcheng Laboratory, Shenzhen 518000, China (e-mail: ft\_liuxiaoqing@mail.scut.edu.cn).}
\thanks{Huanqiang Zeng is with School of Optoelectronic and Communication Engineering, Xiamen University of Technology, Xiamen 361024, China, and also with School of Engineering, Huaqiao University, Quanzhou 362021, China (e-mail: zeng0043@hqu.edu.cn)}}

% \markboth{Journal of \LaTeX\ Class Files,~Vol.~14, No.~8, Aug1ust~2021}%


\maketitle

\begin{abstract}
Multimodal learning aims to integrate information from multiple modalities, however it faces the challenge of modality imbalance, where the dominant modality often suppresses the optimization of weaker modalities due to inconsistent convergence rates.
Existing methods predominantly rely on static modulation or heuristic rules, overlooking significant sample-level distributional variations in the prediction bias between unimodal branches.
Specifically, these methods fail to effectively distinguish outlier samples where the modality gap is exacerbated by low data quality. 
In this paper, we propose a novel framework designed to quantitatively diagnose and dynamically mitigate this imbalance at the sample level. First, we introduce a metric termed Modality Gap to quantify prediction discrepancies between modalities. Through modeling analysis, we observe that the modality gap exhibits a distinct bimodal distribution, unveiling the natural coexistence of modality-balanced and modality-imbalanced sample subgroups within multimodal datasets. Building on this finding, we employ a Gaussian Mixture Model (GMM) to explicitly model this distribution. By leveraging the GMM fit as a prior, we utilize Bayesian posterior probabilities to achieve a probabilistic soft separation of sample subgroups. Finally, we construct a two-stage training framework comprising a Warm-up stage and an Adaptive Training stage. In the core Adaptive Training stage, we employ a GMM-guided Adaptive Loss to dynamically reallocate optimization priorities: it imposes stronger modality alignment penalties on imbalanced samples to rectify bias, while prioritizing multimodal fusion for balanced samples to maximize the exploitation of cross-modal complementary information. Experimental results demonstrate that our method significantly outperforms current state-of-the-art (SOTA) baselines. Furthermore, we demonstrate that fine-tuning on a high-quality balanced subset filtered by the GMM serves as an effective data purification.
%; this approach yields substantial performance gains by eliminating extreme noisy samples that severely hinder learning, even without the adaptive loss.
\end{abstract}

\section{Introduction}
\IEEEPARstart{H}{umans} perceive the world through the integration of multiple senses, where multimodal perception provides more comprehensive information by capturing different aspects of the environment~\cite{baltruvsaitis2018multimodal,liu2025enhanced}. Inspired by human multisensory integration capabilities, multimodal data collected from different sensors has garnered increasing attention in machine learning. In recent years, multimodal learning has demonstrated significant advantages in enhancing the performance of traditional unimodal tasks and addressing new challenging problems, including video classification~\cite{jiang2018modeling,simonyan2014two}, action recognition~\cite{imran2020evaluating}, and audio-visual speech recognition~\cite{czyzewski2017audio}.

\begin{figure}
	\centering
	\includegraphics[width=8.5cm]{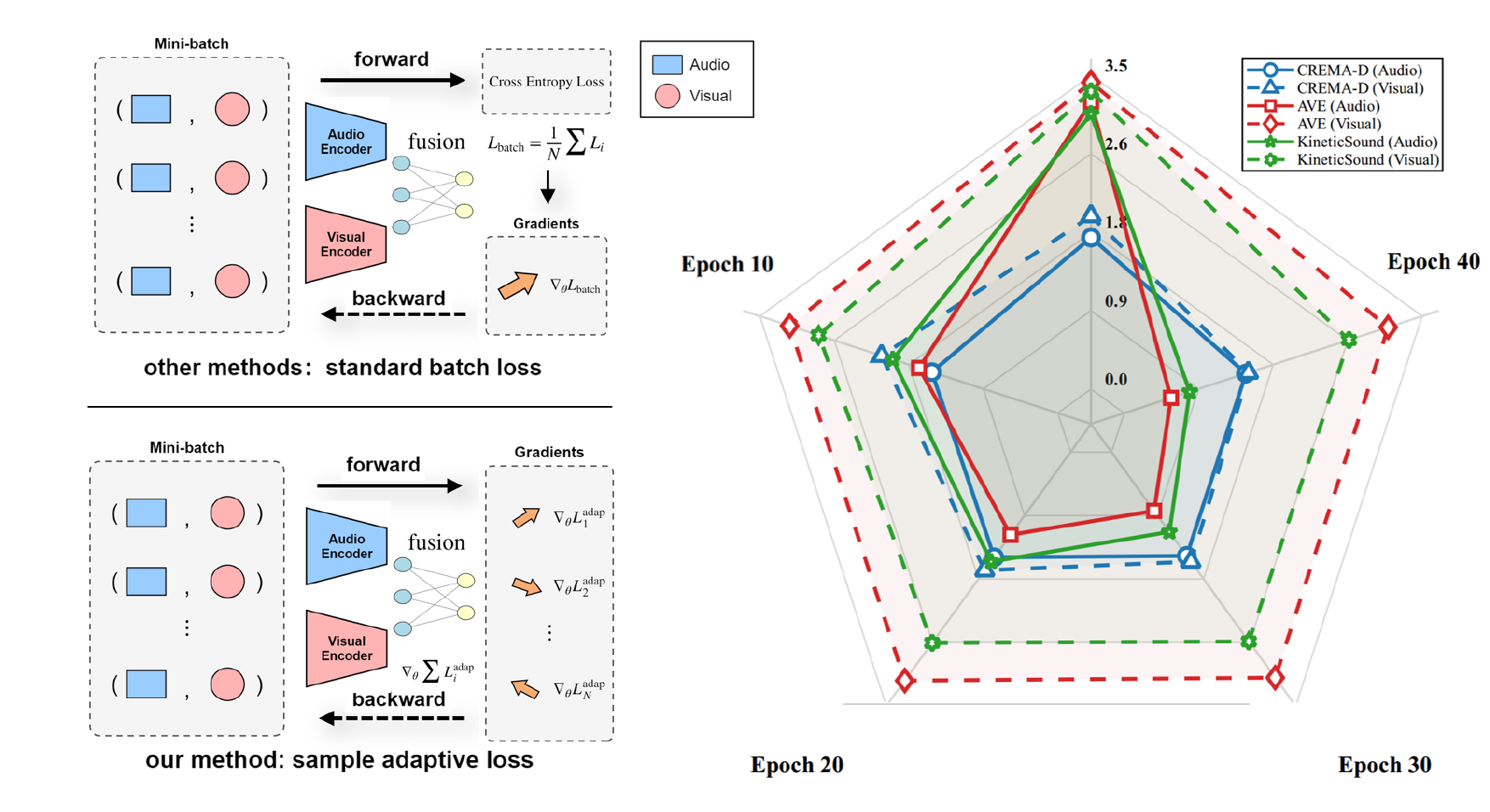}
	\caption{Loss evolution of unimodal branches in OGM-GE~\cite{peng2022balanced}.}
	\vspace{-0.5em}
	\label{intro}
\end{figure}

Compared with unimodal methods, multimodal data generally provide richer and more comprehensive perspectives. Hence, intuitively, multimodal learning should at least match or even surpass the performance of unimodal approaches. However, recent studies have revealed that, in certain cases, multimodal models trained under joint training schemes with a unified learning objective can underperform their unimodal counterparts~\cite{wang2020makes}.
This counterintuitive result arises because different modalities vary in information density and feature extraction difficulty, leading to discrepancies in convergence speed. As a result, the optimization process becomes modality-imbalanced: during the early stages of training, the model tends to overfit the strong modality while the weak modality remains underfitted. The imbalance severely limits the potential and upper bound of overall model performance.

As illustrated in Fig.~\ref{intro}, we employed the method proposed in OGM-GE~\cite{peng2022balanced} to monitor the loss evolution of unimodal branches during training. On the AVE and KineticSound datasets, a significant discrepancy exists between the unimodal losses. Specifically, the visual modality loss remains persistently high, whereas the audio modality loss gradually converges. This suggests that representation learning for the weaker modality stagnates prematurely, preventing the model from effectively exploiting the modality-specific complementary information embedded within it. This phenomenon primarily stems from the fact that most existing methods rely on a standard batch-level loss. In contrast, our approach computes the loss on a per-sample basis, enabling sample-wise differentiated gradient assignment.

To address the multimodal imbalance issue, several studies have emerged recently, leading to significant progress in tackling the challenge of modality imbalance during training~\cite{peng2022balanced, fan2023pmr, xu2023mmcosine, li2023boosting, wei2024diagnosing, sun2024muti, huang2023dominant, su2023utilizing}. Some methods focus on optimizing modality fusion during the feedforward stage to reduce modality imbalance. Greedy~\cite{wu2022characterizing} enhances weak modality learning within the fusion layer to counteract the dominance of stronger modalities. MLA~\cite{zhang2024multimodal} uses an alternating unimodal learning strategy to alleviate mutual interference among modalities during joint optimization, improving joint representation. While these methods reduce modality conflicts, feedforward learning struggles to overcome inertia and often fails to extract meaningful semantic representations from difficult modalities in early training.
To better utilize all modalities, other studies focus on adjusting gradients during backpropagation for balanced multimodal learning. AGM~\cite{li2023boosting} and OGM-GE~\cite{peng2022balanced} regulate gradient updates of modality encoders during backpropagation, suppressing dominant modalities and providing weaker ones with better optimization opportunities. Xu et al.~\cite{wu2022characterizing} proposed MMCosine to improve the objective function by introducing cosine loss via L2 normalization, constraining feature vectors to a shared manifold and reducing imbalance and noise interference caused by varying semantic space magnitudes.
However, existing studies still lack quantitative analysis of modality imbalance and fail to effectively identify balanced versus imbalanced samples. Imbalanced samples often exhibit modality prediction bias, introducing significant data noise during modality fusion and leading to gradient conflicts that hinder joint optimization.

To jointly address the challenges of imbalanced sample noise and optimization trade-offs, we propose an innovative sample-level adaptive optimization framework. During the warm-up phase, we precisely model the degree of imbalance among cross-modal training samples and employ a Gaussian Mixture Model (GMM) to separate modality-balanced and modality-imbalanced subgroups. In the adaptive training phase, the posterior probabilities derived from the GMM are integrated into the loss function as dynamic weights, enabling fine-grained sample-level optimization.
Specifically, we introduce a modality gap metric module (MGM), where the modality gap is defined to quantify the degree of modality balance. After the warm-up phase, we model the distribution of the modality gap using a Gaussian Mixture Model and employ it as a prior for Bayesian posterior inference. This process probabilistically separates naturally existing balanced and imbalanced subgroups within multimodal datasets, establishing a solid foundation for subsequent adaptive optimization.
Furthermore, in the adaptive training stage, recognizing that static strategies struggle to reconcile the optimization trade-off between balanced and imbalanced samples, we design a sample-level adaptive loss function. By softly partitioning samples based on their cross-modal balance, the unified optimization objective is refined into a sample-level dynamic weighted loss, effectively mitigating the imbalance in optimization emphasis between balanced and imbalanced samples.
Finally, to ensure consistency between the optimization strategy and the evolving modality gap distribution during training, we adopt a periodically alternating iterative scheme that jointly updates the GMM-based probabilistic model and network parameters. Additionally, an annealing coefficient is introduced to dynamically adjust the optimization focus—allowing the model to emphasize modality imbalance correction during the early phase of adaptive training and to smoothly transition toward the primary task optimization in later stages.
Our main contributions are summarized as follows:
\begin{itemize}
\item We propose a modality gap metric to quantitatively assess the degree of sample-level modality imbalance, and for the first time model its distribution using a Gaussian Mixture Model (GMM).
\item To simultaneously address the learning differences between modality-balanced and modality-imbalanced samples, we design a sample-level loss function driven by the cross-modal balance distribution, and achieve stable optimization through a two-stage adaptive training scheme.
\item The GMM-based high-quality balanced subset selection serves as an effective data purification strategy, substantially enhancing the performance of baseline models. Extensive experiments on multiple benchmark datasets demonstrate that our proposed method achieves SOTA results, validating the effectiveness and superiority of the proposed approach.
\vspace{0.2em}
\end{itemize}

\section{Relate Work}
\subsection{Multimodal learning}
Multimodal learning aims to construct models capable of processing and associating information from diverse sources, such as text, images, audio, and video. Since real-world information is inherently multimodal, different modalities offer complementary semantic cues; fusing them yields a more robust and comprehensive understanding compared to unimodal approaches. Leveraging these strengths, multimodal learning has achieved remarkable success across various domains. For instance, Visual Question Answering (VQA)~\cite{antol2015vqa,ilievski2017multimodal} facilitates cross-modal reasoning via semantic alignment; Action Recognition~\cite{nagrani2020speech2action,gao2020listen,kazakos2019epic,li2025simultaneous} and Audio-Visual Speech Recognition (AVSR)~\cite{potamianos2004audio,hu2016temporal,yeo2024akvsr} integrate complementary modal information to enhance recognition accuracy; and Multimodal Retrieval~\cite{lin2024cross,liu2025improved,song2024deep} establishes a common semantic space to enable cross-modal search.

\subsection{Imbalanced multimodal learning}
In the realm of multimodal learning, the issue of imbalance presents novel complexities. Beyond traditional class imbalance, multimodal imbalance may also arise, encompassing scenarios such as missing modalities~\cite{yi2024variational,jiang2025boosting} or instances where the quality of one modality is significantly inferior to others (e.g., blurred images or noisy audio). Unlike label or data distribution skew, this form of imbalance refers to the disparity in confidence or contribution among different modalities when predicting the target label. The distribution of this disparity exhibits a distinct bimodal characteristic, suggesting that samples naturally bifurcate into two categories: modality-balanced and modality-unbalanced.

Current approaches addressing multimodal imbalance can be broadly categorized into three streams: feedforward-based, optimization-based, and objective-based methods. Regarding feedforward strategies, OPM~\cite{wei2024fly} mitigates imbalance by discarding features from the dominant modality, while AMCo~\cite{zhou2023adaptive} employs adaptive masking techniques. Optimization-based methods primarily involve gradient modulation strategies~\cite{hua2024reconboost,peng2022balanced},additionally, PMR~\cite{fan2023pmr} adjusts gradient magnitudes based on class prototypes to accelerate the learning of weaker modalities. Finally, objective-based approaches tackle the problem by modifying the training objective functions~\cite{xu2023mmcosine, du2023uni, wang2020makes, wei2024mmpareto, yang2024facilitating}.

\section{Proposed Method}
\subsection{Framework and Notations}
\label{Framework_and_notations}
\begin{figure*}
	\centering
	\includegraphics[width=0.9\linewidth]{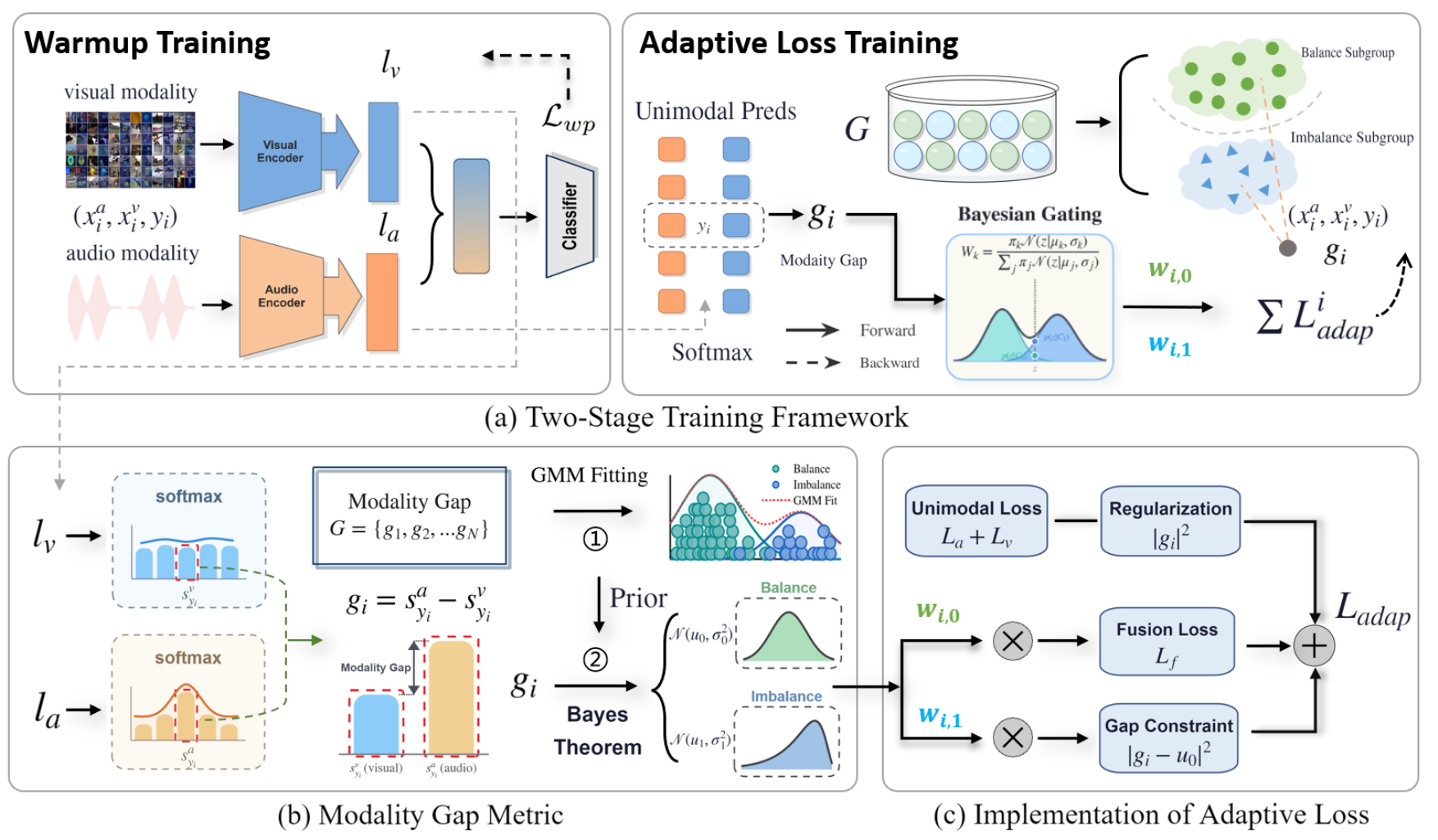}
	\caption{The proposed two-stage training framework: Warm-up and Adaptive Loss Training.}
	\vspace{-0.5em}
	\label{pipeline}
\end{figure*}
As illustrated in Fig.~\ref{pipeline}, we design a two-stage training framework. The initial warm-up stage aims to guide the model toward preliminary convergence and establish foundational modal representations. The subsequent adaptive training stage constitutes the core of this work, by employing an alternating iterative strategy, it synergizes prior modeling based on Gaussian Mixture Model (GMM) with sample-level adaptive optimization.

We formulate the audio–visual classification task as a supervised learning problem. Let the dataset be $\mathcal{D}=\{x_{i}, y_{i}\}_{i=1}^{N}$, where each input $x_{i}=(x^{a}_{i}, x^{v}_{i})$ contains paired audio and visual modalities, and $y_i \in \{1,2,\ldots,M\}$ denotes the ground-truth label among $M$ classes. We employ two independent single-modality backbone networks, $\varphi^{a}(\theta^{a},\cdot)$ and $\varphi^{v}(\theta^{v},\cdot)$, to extract features from the audio and visual streams, respectively, with $\theta^{a}$ and $\theta^{v}$ representing their encoder parameters. We adopt simple feature concatenation as the fusion strategy, after which the concatenated representation is passed through a fully connected classification head, with weight matrix $W \in \mathbb{R}^{M \times (d_{\varphi_a}+d_{\varphi_v})}$ and bias $b \in \mathbb{R}^{M}$, to yield the final multimodal fusion logit vector $f(x_i)$, which is defined as:
\begin{equation}
\label{output}
f(x_{i})=W[\varphi^{a}(\theta^{a},x^{a}_{i});\varphi^{v}(\theta^{v},x^{v}_{i})]+b.
\end{equation}

To quantitatively isolate the independent contributions of each modality in joint prediction, we leverage the linearity of matrix multiplication and conceptually partition the weight matrix $W$ into two submatrices, $W^a$ and $W^v$. Consequently, Eq.~\ref{output} can be reformulated as the sum of the two modality-specific logit components:
\begin{equation}
	\label{output_re} 
	f(x_{i}) = \underbrace{W^{a}\varphi^{a}(x^{a}_{i}) + \frac{b}{2}}_{l_i^a} + \underbrace{W^{v}\varphi^{v}(x^{v}_{i}) + \frac{b}{2}}_{l_i^v},
\end{equation}
where $l_i^a$ and $l_i^v$ represent the logits attributed to the audio and visual modalities, respectively. For consistency of scale, the bias term $b$ is evenly distributed across the two components.

The predicted probability distributions for both the fusion and unimodal branches are derived via the Softmax function, as follows:
\begin{equation}
	\label{unimodal-softmax-m}
	s^{(m)} = \operatorname{softmax}\left(l^{(m)}\right), \quad m \in \{f,a,v\},
\end{equation}
the probability allocated to the target class for the $i$-th sample, denoted as $s^{(m)}_{i, y_i}$ where $m \in \{f,a,v\}$, reflects the prediction confidence of the respective branch. Specifically, a value of $s^{(m)}_{i, y_i}$ closer to 1 indicates higher discriminative power of the features within that branch. We employ the standard Cross-Entropy (CE) loss for both the primary multimodal task and the auxiliary unimodal tasks, expressing them via a unified formulation:
\begin{equation}
	\label{L_CE}
	\begin{aligned}
    \mathcal{L}_{m} = \frac{1}{N} \sum_{i=1}^{N} \text{CE}(y_i, s_i^{(m)}), \quad m\in\{f,v,a\}.
\end{aligned}
\end{equation}

\subsection{Modality Balance Metric}
To quantify the degree of multimodal imbalance at the sample level, we propose a modality gap metric, denoted as $g_i$, which is defined from two complementary perspectives: confidence and information entropy.

Intuitively, the relative strength between modalities is directly reflected in the prediction confidence of each modality branch for the ground-truth class. Specifically, $s_{i, y_i}^a$ and $s_{i, y_i}^v$ denote the predicted probabilities assigned by the audio and visual branches, respectively, to the true label $y_i$ for the $i$-th sample. Assuming that the audio modality is dominant (a phenomenon widely observed in the target dataset), we define the modality gap as the difference between the predicted probabilities of the two modalities on the correct class:
\begin{equation}
	\label{modality-gap-1}
	\begin{aligned}
		g_i = s_{i, y_i}^a - s_{i, y_i}^v,
	\end{aligned}
\end{equation}
the positive $g_i$ indicates that the audio modality dominates the prediction, whereas a value near zero implies comparable contributions from both modalities, signifying a balanced state.

However, confidence-based metrics essentially constitute ground-truth-dependent point estimates, which are not only sensitive to numerical fluctuations but also fail to capture the global characteristics of the predictive distribution. To overcome this limitation, we formulate a more robust modality gap measure from an information-theoretic perspective. Specifically, we introduce the Kullback–Leibler (KL) divergence between the predicted distribution and the uniform distribution $\mathcal{U}$ (where $u_j = 1/M$). The divergence $D_{KL}(s^{(m)} | \mathcal{U})$ quantifies the informativeness and certainty of a given modality’s output. A larger divergence corresponds to a sharper and more confident predictive distribution.
\begin{equation}
	\label{modality-gap-2}
	\begin{aligned}
		g_i = D_{KL}(s_i^a \| \mathcal{U}) - D_{KL}(s_i^v \| \mathcal{U}),
	\end{aligned}
\end{equation}
the Kullback–Leibler (KL) divergence between the predicted distribution $s_i^{(m)}$ and the uniform distribution $\mathcal{U}$ is given by:
\begin{equation}
	\label{modality-gap-KL-cal}
	\begin{aligned}
		D_{KL}(s_i^{(m)} \| \mathcal{U}) = \sum_{j=1}^{M} s_{i,j}^{(m)} \log\left(\frac{s_{i,j}^{(m)}}{1/M}\right),\quad m\in\{v,a\}.
	\end{aligned}
\end{equation}

Given that the raw $g_i$ values derived from Eq.~\ref{modality-gap-1} and Eq.~\ref{modality-gap-2} differ in scale and range, we apply Z-score normalization to standardize the metric distribution, thereby ensuring a consistent input space for subsequent probability modeling.
\begin{equation}
	\label{modality-gap-norm}
	\begin{aligned}
		\hat{g}_i = (g_i - \mu_{\mathcal{G}}) / \sigma_{\mathcal{G}},
	\end{aligned}
\end{equation}
where $\mu_{\mathcal{G}}$ and $\sigma_{\mathcal{G}}$ denote the mean and standard deviation of the set $\mathcal{G}$, respectively.

\begin{figure}[ht] % 使用 [ht] 尝试放置在顶部或当前位置
	\centering
	\includegraphics[width=6.5cm]{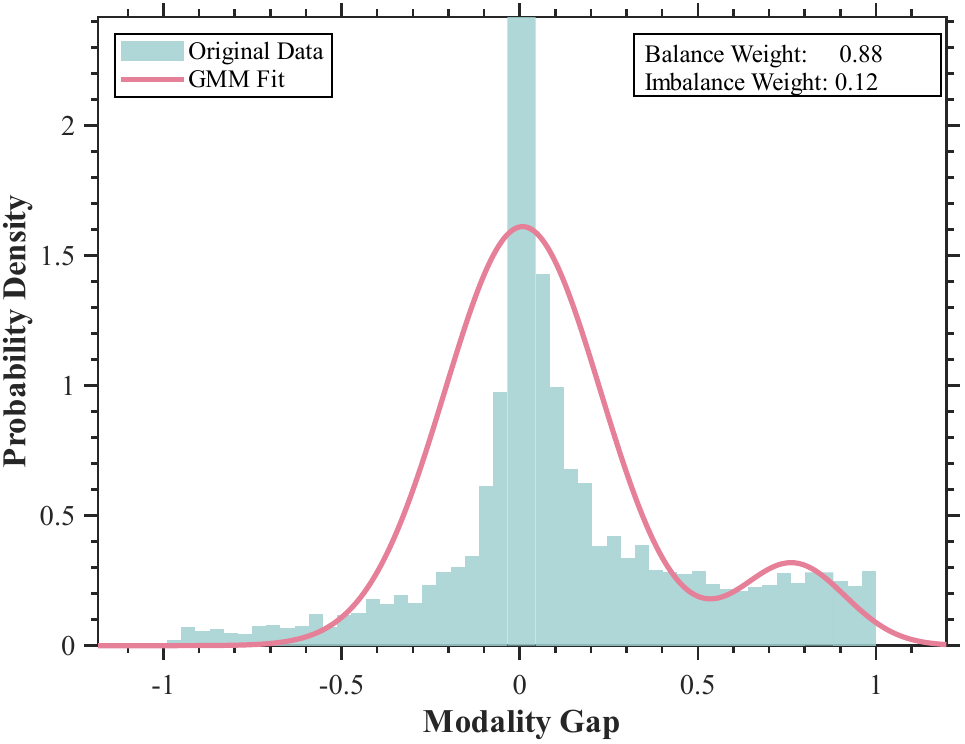} 
	\vspace{-0.2em}
	\caption{Visualization of the GMM fitting of modality gap($\mathbf{g}$) on the CREMA-D Dataset.}
	\label{CREMAD_GMM_modality_gap_step_0}
	\vspace{-0.5em}
\end{figure}

After the warm-up training stage, we leverage the model’s acquired feature extraction capability together with the memorization effect of neural networks to aggregate the modality gaps of all training samples. The aggregated gaps are further normalized to construct the set $\mathcal{G} = \{\hat{g}_1, \hat{g}_2, \ldots, \hat{g}_N\}$. As illustrated in Fig.~\ref{CREMAD_GMM_modality_gap_step_0}, the visualization reveals a pronounced bimodal structure, indicating that the multimodal dataset naturally consists of two latent subgroups. The primary peak is concentrated around zero, corresponding to modality-balanced samples. These samples exhibit consistent prediction confidence across modalities, thereby facilitating effective multimodal fusion. In contrast to a standard normal distribution, the right tail does not decay rapidly but instead forms a secondary mode. This secondary mode corresponds to a relatively small yet non-negligible multimodal imbalance subgroup. Samples in this subgroup exhibit substantial discrepancies between modality-specific predictions, which serve as a major source of gradient conflicts and optimization instability during training.

Given the substantial data noise introduced by the multimodal imbalance subgroup, this motivates us to perform probabilistic modeling to achieve a soft separation of sample subgroups, which serves as the basis for subsequent adaptive optimization. Moreover, since the bimodal characteristic of the modality gap distribution becomes evident after the warm-up stage, we naturally adopt a two-component Gaussian Mixture Model (GMM) to explicitly model the set $\mathcal{G}$. The probability density function of the resulting prior distribution over the modality gap is formulated as follows:
\begin{equation} 
	\label{GM-fitting-prob-density} 
	\begin{aligned} 
		p(g | \Theta) = \pi_{0} \mathcal{N}(g | \mu_0, \sigma_0^2) + \pi_{1} \mathcal{N}(g | \mu_1, \sigma_1^2), 
	\end{aligned} 
\end{equation}
where $\Theta = \{\pi_k, \mu_k, \sigma_k\}_{k \in \{0,1\}}$ denotes the parameter set of the GMM, where $\pi_k, \mu_k$, and $\sigma_k$ represent the mixture weight, mean, and variance of the $k$-th Gaussian component, respectively. Based on the distributional characteristics, we identify the primary peak centered near zero ($k=0$) as the modality-balanced distribution, and the long-tail component (secondary peak) deviating from zero ($k=1$) as the distribution corresponding to multimodal imbalance. 

To map this distribution-level prior onto individual samples, we compute the probability that a given sample belongs to each latent subgroup. Formally, we introduce a latent variable $z_i \in \{0, 1\}$ to denote the subgroup membership of sample $i$. According to the definition of the Gaussian Mixture Model (GMM), the prior probability that sample $i$ belongs to the $k$-th subgroup, $P(z_i = k)$, is given by:
\begin{equation} 
	\label{GM-fitting-prior} 
	\begin{aligned} 
        P(z_i = k) = \pi_k,
	\end{aligned} 
\end{equation}
similarly, the conditional probability density of $\hat{g}_i$ given $z_i = k$, denoted as $p(\hat{g}_i \mid z_i = k)$, is defined as:
\begin{equation} 
	\label{GM-fitting-condition-prob} 
	\begin{aligned} 
        p(\hat{g}_i | z_i = k) = \mathcal{N}(\hat{g}_i | \mu_k, \sigma_k^2).
	\end{aligned} 
\end{equation}

Consequently, according to the law of total probability, the marginal probability density of $\hat{g}_i$ is given by:
\begin{equation} 
	\label{GM-fitting-marginal-prob} 
	\begin{aligned} 
        p(\hat{g}_i) = \sum_{j=0}^{1} \pi_{j} \mathcal{N}(\hat{g}_i | \mu_j, \sigma_j^2).
	\end{aligned} 
\end{equation}

Based on the above formulations, we apply Bayes' Theorem to compute the posterior probability that sample $i$ belongs to the $k$-th subgroup, denoted as $w_{i,k} = P(z_i = k \mid \hat{g}_i)$:
\begin{equation} 
	\label{GM-fitting-bayes} 
	\begin{aligned} 
		w_{i,k} = \frac{P(z_i = k) p(\hat{g}_i | z_i = k)}{p(\hat{g}_i)}.
	\end{aligned} 
\end{equation}

By explicitly expanding the above expression, we obtain the posterior probability that sample $i$ belongs to the modality-balanced subgroup ($k = 0$), denoted as $w_{i,0}$:
\begin{equation} 
	\label{GM-fitting-w0} 
	\begin{aligned} 
		w_{i,0} = \frac{\pi_{0} \mathcal{N}(\hat{g}_i | \mu_0, \sigma_0^2)}{\pi_{0} \mathcal{N}(\hat{g}_i | \mu_0, \sigma_0^2) + \pi_{1} \mathcal{N}(\hat{g}_i | \mu_1, \sigma_1^2)},
	\end{aligned} 
\end{equation}
similarly, we obtain the posterior probability that sample $i$ belongs to the multimodal imbalance subgroup ($k = 1$), denoted as $w_{i,1}$.
\begin{equation} 
	\label{GM-fitting-w1} 
	\begin{aligned} 
		w_{i,1} = \frac{\pi_{1} \mathcal{N}(\hat{g}_i | \mu_1, \sigma_1^2)}{\pi_{0} \mathcal{N}(\hat{g}_i | \mu_0, \sigma_0^2) + \pi_{1} \mathcal{N}(\hat{g}_i | \mu_1, \sigma_1^2)},
	\end{aligned} 
\end{equation}
$w_{i,0}$ and $w_{i,1}$ constitute probabilistic measures that characterize the degree of sample-level multimodal imbalance. These posterior probabilities are employed as dynamic soft weights in the subsequent adaptive training stage, guiding the adaptive reweighting of the loss function.

\subsection{Two-Stage Adaptive Optimization}
\label{adaptive-training}
Obtaining a reliable statistical distribution of the modality gap serves as a prerequisite for achieving probabilistic sample separation and facilitating subsequent adaptive optimization. To this end, we design a two-stage training framework.

Warm-up training constitutes the initial phase of our two-stage adaptive framework, primarily designed to steer the model toward preliminary convergence. The objective of this phase is to learn basic multimodal fusion while simultaneously maintaining strong unimodal features. This establishes reliable logit distributions requisite for modeling the modality gap in the second stage. To this end, we incorporate a unimodal loss into the objective function of this phase as an auxiliary term:
\begin{equation}
	\label{warm-up-loss}
	\begin{aligned}
		\mathcal{L}_{wp} = \mathcal{L}_{f} + \mathcal{L}_a + \mathcal{L}_v.
	\end{aligned}
\end{equation}

In the second-stage adaptive training, we employ the GMM as a probabilistic modeling framework and apply Bayes’ theorem to compute the posterior probabilities that each sample belongs to the balanced subgroup ($w_{i,0}$) and the imbalance subgroup ($w_{i,1}$). Based on this probabilistic measure of sample-level imbalance, we replace the static optimization objective with a sample-wise differentiated optimization strategy that accommodates the distinct learning requirements of balanced and imbalanced samples. For samples with high $w_{i,0}$, indicating modality balance, we impose stronger fusion penalty to fully exploit complementary cross-modal information. In contrast, for samples with high $w_{i,1}$, indicating multimodal imbalance, we impose stronger explicit geometric constraints to narrow their modality gap, thereby encouraging the imbalanced distribution to shift toward the center of the balanced distribution ($\mu_0$). Furthermore, a regularization term ($\lambda |g_i|^2$) is introduced for all samples to penalize excessively large modality gaps. By converting the GMM posterior probabilities into dynamic weights, we formulate the following sample-level adaptive loss function $\mathcal{L}^i_{adap}$:
\begin{equation}
	\label{adaptive-loss}
	\begin{aligned}
		\begin{split}
	\mathcal{L}^i_{adap}=\alpha \cdot w_{i,0} \cdot \mathcal{L}_{f} +  \lambda_t \cdot (\beta \cdot |g_i|^2 \\+ \gamma \cdot w_{i,1} \cdot |g_i-u_0|^2 + \mathcal{L}_a + \mathcal{L}_v),
	\end{split}
	\end{aligned}
\end{equation}
where $\lambda_t = \rho^t$ serves as the decay factor at epoch $t$, with the decay rate $\rho \in (0, 1)$. During the initial training phase, the weight assigned to the modality gap penalty is substantial, compelling the model to rapidly reduce the modality gap. As training progresses, the overall modality gap diminishes and its distribution tends towards unimodality, concurrently, the probability of samples being identified as unbalanced by the GMM decreases. Correspondingly, $\lambda_t$ decays, signifying a gradual relaxation of the imbalance penalty, thereby allowing the model to shift its focus towards the nuances of the primary classification task. Given that the modality gap evolves dynamically during training, we employ an alternating iterative strategy to coordinate the GMM fitting and the adaptive training process. 

\subsection{Theoretical Analysis}
To thoroughly investigate the specific mechanism of the adaptive loss function (Eq.~\ref{adaptive-loss}) from the perspective of optimization dynamics, we isolate the component that directly constrains the modality gap, along with its associated regularization term, for independent analysis. These components constitute the optimization driving force for correcting multimodal imbalance and is defined as follows:
\begin{equation}
	\label{L-gap}
	\begin{aligned}
		\begin{split}
    \mathcal{L}_{gap}^i = \gamma \cdot w_{i,1} \cdot |g_i - u_0|^2 + \beta \cdot |g_i|^2,
	\end{split}
	\end{aligned}
\end{equation}
incorporating the definition of the modality gap (Eq.~\ref{modality-gap-1}), we reformulate $\mathcal{L}_{gap}^i$ as a function of the modal prediction probabilities $s_i^a$ and $s_i^v$, as follows:
\begin{equation}
	\label{L-gap-1}
	\begin{aligned}
		\begin{split}
    \mathcal{L}_{gap}^i = \gamma \cdot w_{i,1} \cdot |s_i^a - s_i^v - u_0|^2 + \beta \cdot |s_i^a - s_i^v|^2.
	\end{split}
	\end{aligned}
\end{equation}

Subsequently, to elucidate how this loss function influences model updates via backpropagation, we conduct a gradient analysis focusing on the weak modality (using the visual modality $\theta^v$ as an example). By applying the chain rule, the gradient of this loss term with respect to the visual encoder parameters is derived as follows:
\begin{equation}
	\label{L-gap-grad}
	\begin{split}
		\nabla_{\theta^v} \mathcal{L}_{gap}^i &= -2 \cdot w_{i,1} \cdot (g_i - u_0) \cdot \nabla_{\theta^v} s_i^v - 2 \cdot \beta \cdot g_i \cdot \nabla_{\theta^v} s_i^v \\
		&= -2 \cdot \left[ w_{i,1} \cdot (g_i - u_0) + \beta \cdot g_i \right] \cdot \nabla_{\theta^v} s_i^v.
	\end{split}
\end{equation}
When a sample is identified as imbalanced by the GMM, the weight $w_{i,1}$ becomes significant, and both $g_i$ and the deviation $g_i - u_0$ exhibit large values. Consequently, the gradient magnitude in the aforementioned formulation is significantly amplified. This indicates that imbalanced samples make a substantial gradient contribution to the weak modality; the resulting larger gradient updates facilitate overcoming the optimization stagnation of the weak modality. Conversely, when a sample is identified as balanced, $w_{i,1} \to 0$, and both $g_i$ and $g_i - u_0$ remain small. In this scenario, the optimization objective smoothly transitions towards modal fusion learning, thereby focusing on exploiting complementary information between modalities.

\section{Experiments}
\subsection{Datasets}
\noindent \textbf{CREMA-D}\cite{cao2014crema} is an audio-visual dataset for speech emotion recognition, containing 7,442 video clips of 2-3 seconds from 91 actors speaking several short words. This dataset consists of 6 most usual emotions: \textit{angry}, \textit{happy}, \textit{sad}, \textit{neutral}, \textit{disgust} and \textit{fear}. The whole dataset is randomly divided into 6,698-sample training set and validation set according to the ratio of 9:1, as well as a 744-sample testing set. \\
\noindent \textbf{AVE}\cite{tian2018audio} is an audio-visual video dataset for audio-visual event localization, which covers 28 event classes and consists of 4,143 10-second videos with both auditory and visual tracks as well as frame-level annotations. All videos are collected from YouTube. In experiments, the split of the dataset follows~\cite{tian2018audio}. \\
\noindent \textbf{Kinetics-Sounds} (KS) \cite{arandjelovic2017look} is a dataset containing 31 human action classes selected from Kinetics dataset~\cite{kay2017kinetics} which contains 400 classes of YouTube videos. All videos are manually annotated for human action using Mechanical Turk and cropped to 10 seconds long around the action. The 31 classes were chosen to be potentially manifested visually and aurally, such as playing various instruments. This dataset contains 19k 10-second video clips (15k training, 1.9k validation, 1.9k test).

We formulate the learning objective across these three datasets as a standard supervised classification task. The specific data partition statistics used in our experiments are presented in Table~\ref{tab:dataset_stats}. Notably, the counts of valid samples for AVE and KineticSound listed in the table are slightly lower than the nominal totals reported in the original papers. This minor discrepancy is primarily attributed to the unavailability of certain YouTube source links over time, as well as the exclusion of corrupted samples.
\begin{table}[t]
	\centering
	\caption{Statistics of the datasets used in our experiments.}
	\label{tab:dataset_stats}
	\begin{tabular*}{\linewidth}{l@{\extracolsep{\fill}}cccc}
		\toprule
		\textbf{Dataset} & \textbf{Total} & \textbf{Train} & \textbf{Test} & \textbf{classes} \\
		\midrule
		CREMA-D  & 7442 & 6698 & 744 & 6 \\
		AVE      & 3716 & 3312 & 402 & 28 \\
		KineticSound & 17284 & 14739 & 2545 & 31 \\
		\bottomrule
	\end{tabular*}
\end{table}

\subsection{Experimental settings}
We employ ResNet-18~\cite{he2016deep} as the backbone network and train it from random initialization. The choice of model architecture and initialization strategy follows prior work on multimodal imbalance learning, ensuring consistency and fair comparison. During training, we adopt the stochastic gradient descent (SGD) optimizer with momentum (momentum factor set to 0.9), and the initial learning rate is configured to $2\times10^{-3}$. All models are trained on NVIDIA RTX 3090 (Ti) GPUs.

\subsection{Comparison on the multimodal task}

To comprehensively evaluate the effectiveness of our proposed method, we selected a suite of representative baselines addressing multimodal imbalance, including G-Blending~\cite{wang2020makes}, OGM-GE~\cite{peng2022balanced}, Greedy~\cite{wu2022characterizing}, PMR~\cite{fan2023pmr}, AGM~\cite{li2023boosting}, MLA~\cite{zhang2024multimodal}, D\&R~\cite{wei2024diagnosing}, and OPM\&OGM~\cite{wei2024fly}.

\begin{figure*}[ht]
	\centering
	
	% --- 第一排 ---
	\begin{subfigure}[b]{0.28\textwidth}
		\centering
		\includegraphics[width=\textwidth]{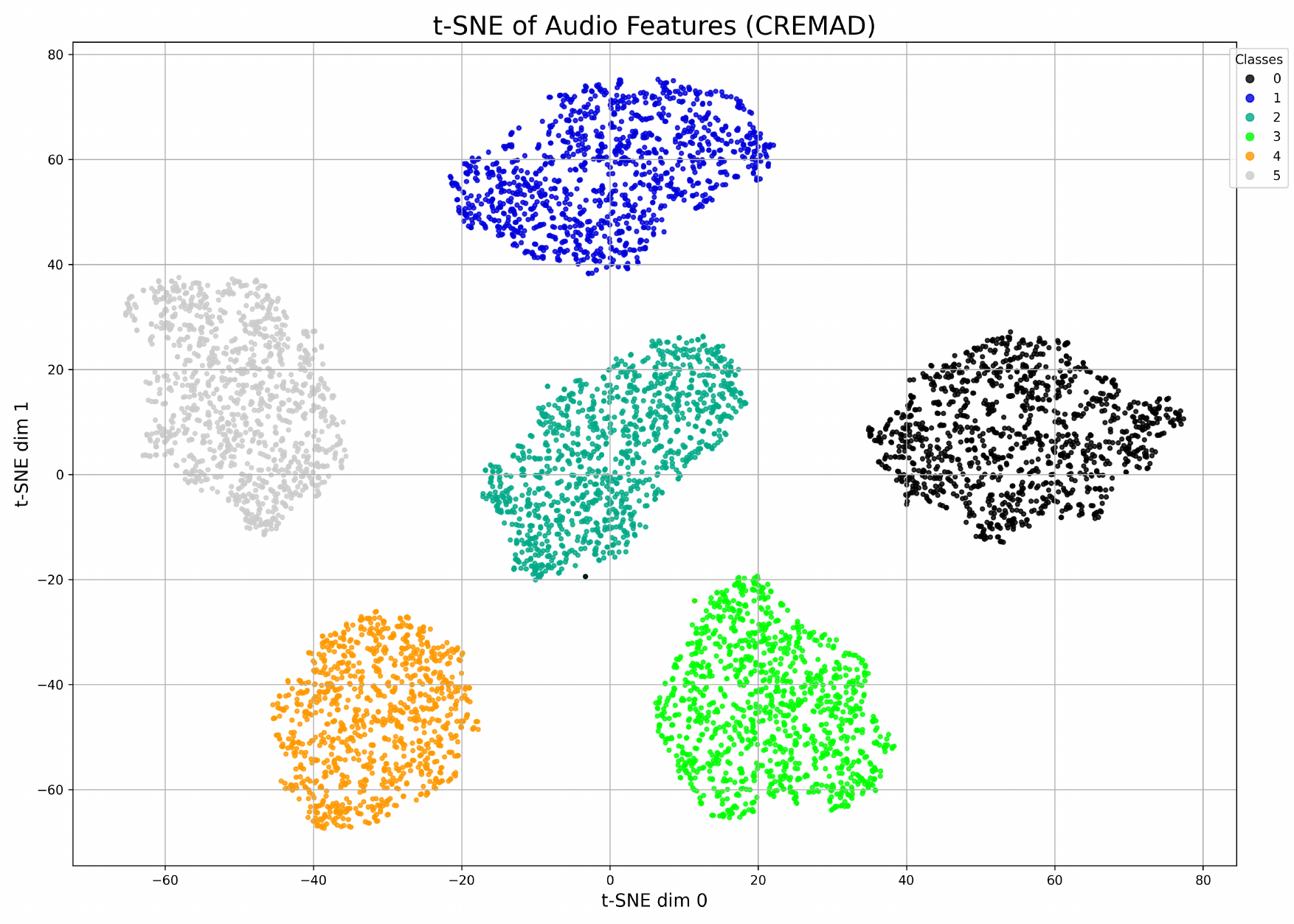}
		\caption{Audio}
	\end{subfigure}
	\hspace{1.2em}
	\begin{subfigure}[b]{0.28\textwidth}
		\centering
		\includegraphics[width=\textwidth]{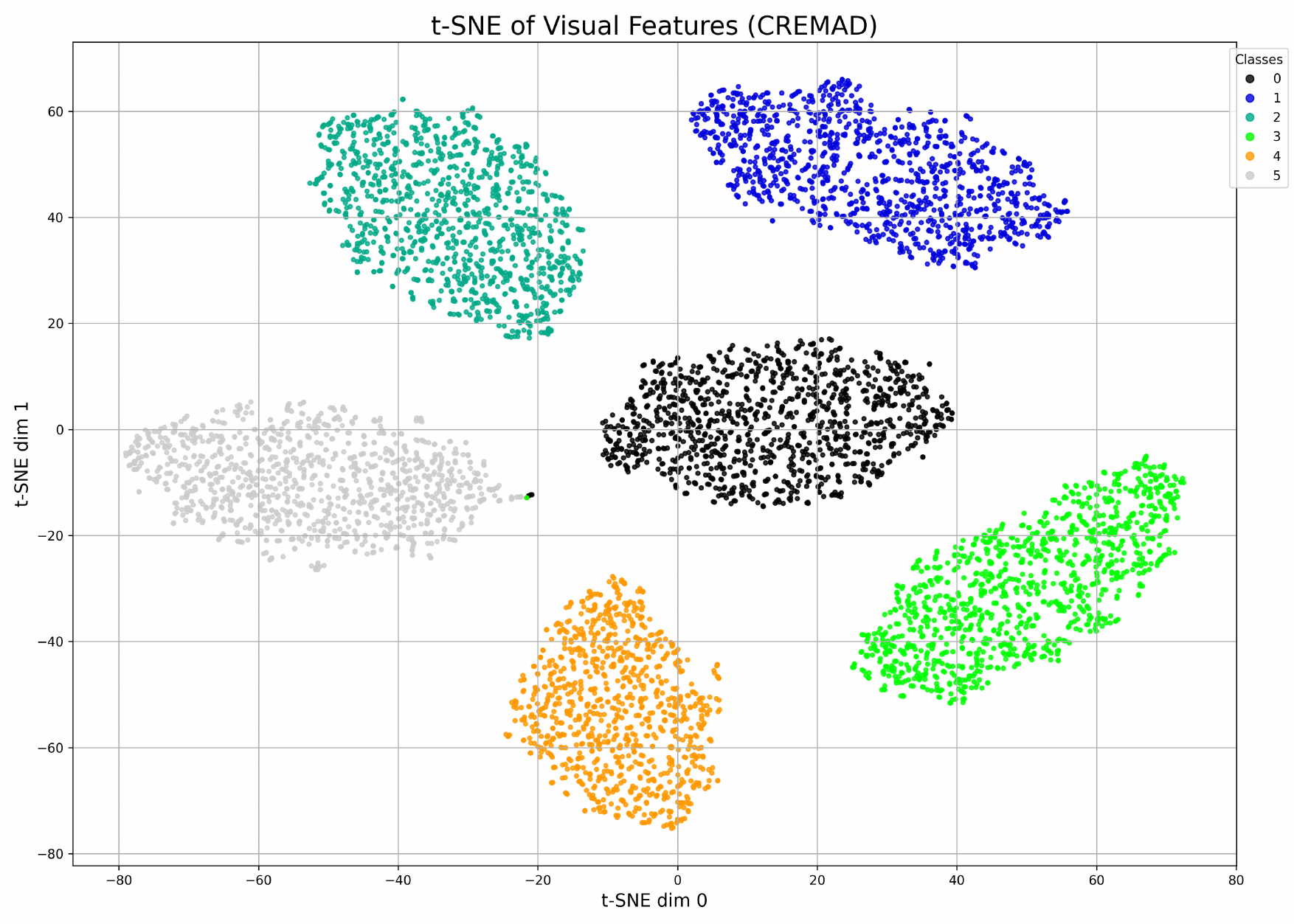}
		\caption{Visual}
	\end{subfigure}
	\hspace{1.2em}
	\begin{subfigure}[b]{0.28\textwidth}
		\centering
		\includegraphics[width=\textwidth]{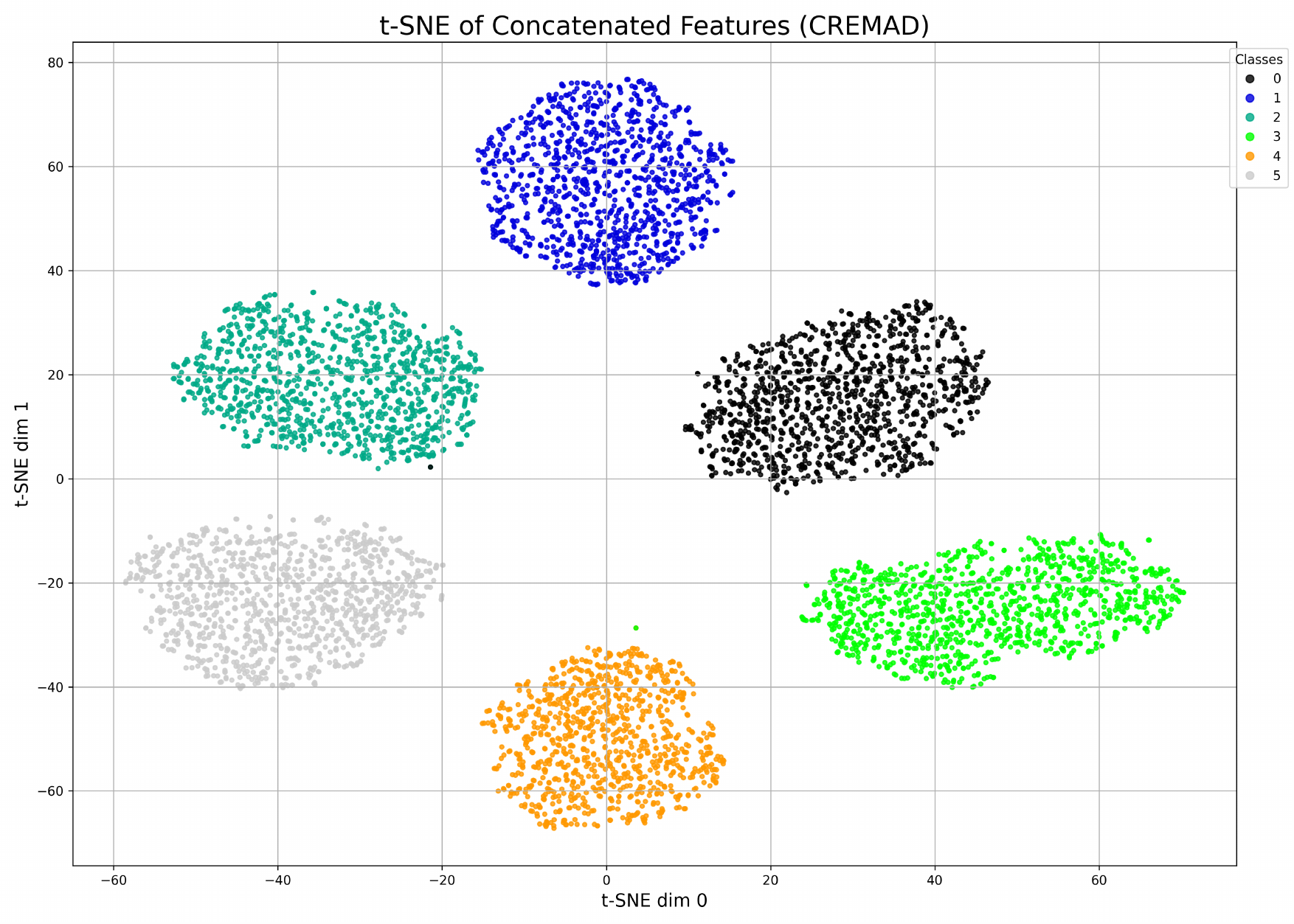}
		\caption{Fusion}
	\end{subfigure}

	% --- 第二排 ---
	\begin{subfigure}[b]{0.28\textwidth}
		\centering
		\includegraphics[width=\textwidth]{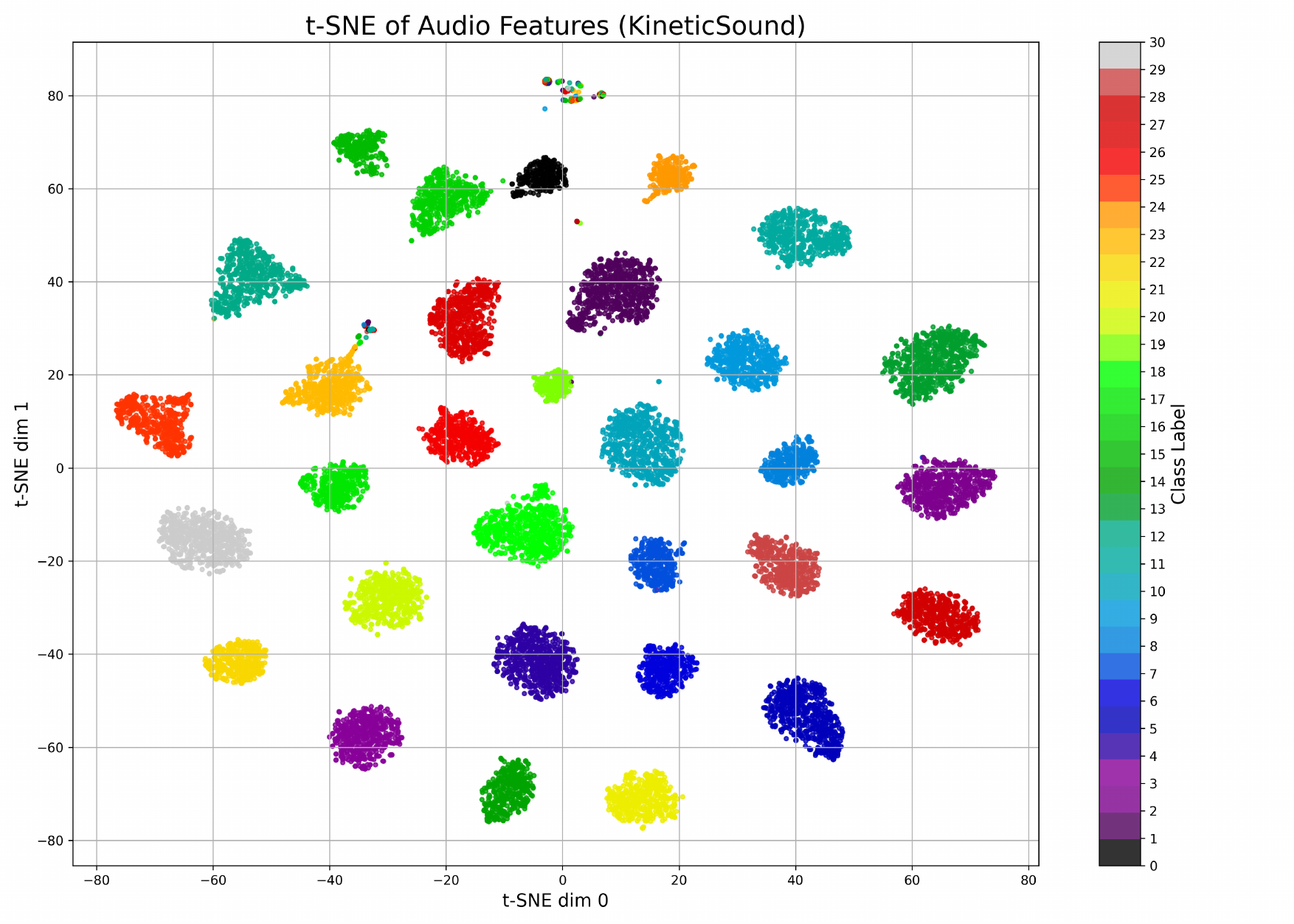}
		\caption{Audio}
	\end{subfigure}
	\hspace{1.2em}
	\begin{subfigure}[b]{0.28\textwidth}
		\centering
		\includegraphics[width=\textwidth]{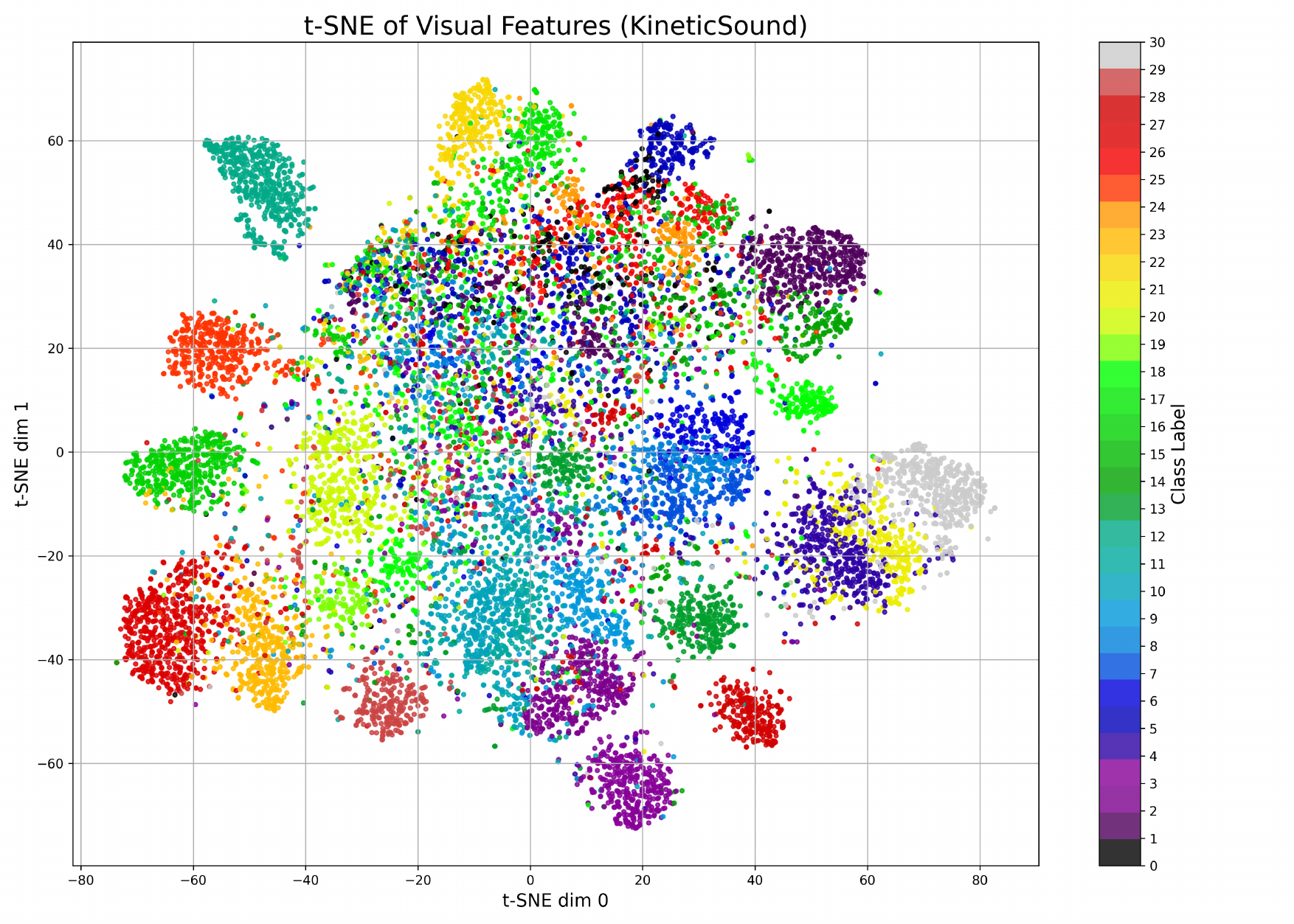}
		\caption{Visual}
	\end{subfigure}
	\hspace{1.2em}
	\begin{subfigure}[b]{0.28\textwidth}
		\centering
		\includegraphics[width=\textwidth]{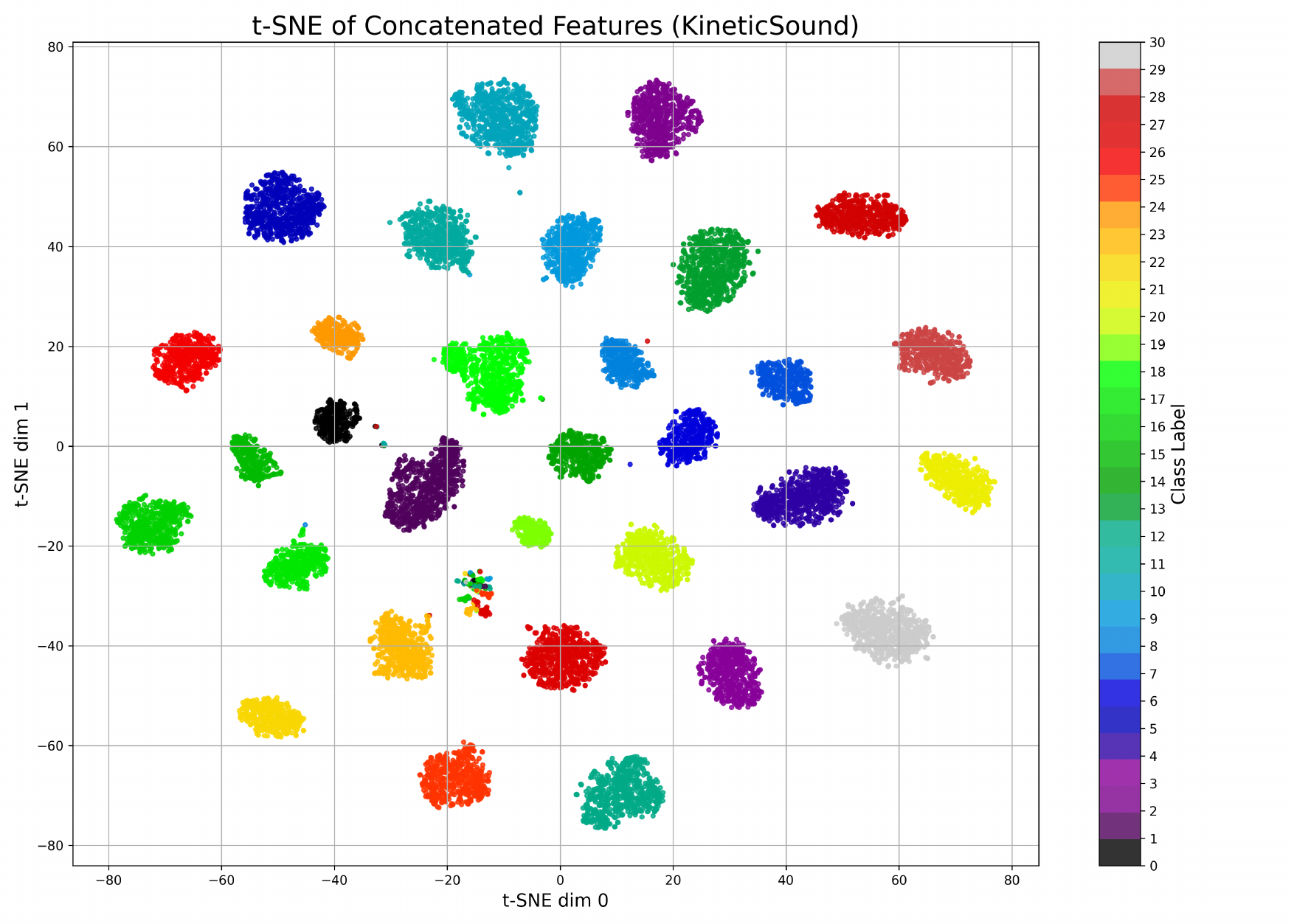}
		\caption{Fusion}
	\end{subfigure}

	\caption{t-SNE Visualization on CREMA-D and KineticSound dataset}
	\label{t-SNE}
	\vspace{-0.5em}
\end{figure*}

As presented in Table~\ref{tab:main_results}, our method exhibits significant performance superiority across all three benchmark datasets. Specifically, it achieves accuracies of 80.65\%, 70.40\%, and 72.61\% on CREMA-D, AVE, and KineticSound, respectively, establishing new state-of-the-art (SOTA) benchmarks. Notably, even the KL divergence-based variant (Ours-KL Gap) outperforms the vast majority of baseline methods. Taken together, these consistent improvements across multiple datasets provide compelling evidence of the effectiveness and generalizability of our approach in addressing multimodal imbalance. Furthermore, to provide an in-depth analysis of feature learning from the perspective of manifold structure, we employed t-SNE \cite{maaten2008visualizing} to visualize both the unimodal features and the fused multimodal features, as illustrated in Fig.~\ref{t-SNE}.

\begin{table*}[t]
	\caption{Comparative experimental results on the \textbf{CREMA-D}, \textbf{AVE}, and \textbf{KineticSound} datasets.}

	\centering
	\begin{tabular*}{\textwidth}{@{\extracolsep{\fill}}lcccccc}
		\toprule[1pt]
		\multirow{2}{*}{Method} 
		& \multicolumn{2}{c}{CREMA-D} 
		& \multicolumn{2}{c}{AVE} 
		& \multicolumn{2}{c}{KineticSound} \\
		\cmidrule(lr){2-3} \cmidrule(lr){4-5} \cmidrule(lr){6-7}
		& Accuracy (\%) & Macro F1 (\%) 
		& Accuracy (\%) & Macro F1 (\%) 
		& Accuracy (\%) & Macro F1 (\%) \\
		\hline
		Baseline (Concat) & 67.47 & 67.80 & 64.68 & 62.24 & 65.54 & 64.52 \\
		G-Blending\cite{wang2020makes} & 69.89 & 70.41 & 66.17 & 62.30 & 68.60 & 68.64 \\
		OGM-GE\cite{peng2022balanced} & 68.95 & 69.39 & 65.67 & 63.00 & 68.88 & 68.10 \\
		PMR\cite{fan2023pmr} & 68.55 & 68.99 & 63.43 & 59.83 & 65.62 & 65.36 \\
		MMCosine\cite{xu2023mmcosine} & 72.45 & 72.57 & 63.18 & 59.87 & 67.50 & 66.66 \\
		AGM\cite{li2023boosting} & 70.16 & 70.67 & 64.92 & 60.19 & 66.50 & 66.49 \\
		MLA\cite{zhang2024multimodal} & \underline{79.70} & \underline{79.94} & 65.92 & 64.91 & \underline{71.35} & \underline{71.23} \\
		D\&R\cite{wei2024diagnosing} & 75.13 & 76.00 & \underline{68.66} & 64.89 & 70.84 & 69.84 \\
		OPM\&OGM\cite{wei2024fly} & 75.10 & 75.91 & 67.41 & 63.46 & 69.00 & 68.11 \\
		\hline
		\textbf{Ours (Prob Gap)} 
		& \textbf{80.65} & \textbf{80.82}
		& \textbf{70.40} & \textbf{66.70}
		& \textbf{72.61} & \textbf{71.64} \\
		\hline 
		\textbf{Ours (KL Gap)} 
		& 79.84 & 80.24
		& 69.65 & 66.51
		& 72.26 & 71.21 \\
		\bottomrule[1pt]
	\end{tabular*}

	\vspace{-0.5em}
	\label{tab:main_results}
\end{table*}

\subsection{Ablation study}

We performed an ablation study on the adaptive loss function (Eq.~\ref{adaptive-loss}) during the adaptive training phase. To evaluate the contribution of individual components, we set the hyperparameters $\alpha=0$, $\beta=0$, and $\gamma=0$ respectively, and excluded the unimodal loss. Furthermore, to validate the efficacy of the GMM-based soft weighting mechanism, we eliminated the weighting effect by setting $w_{i,0}=1$ and $w_{i,1}=1$. The experimental results are summarized in Table~\ref{loss-ablation}. Overall, the Full Model achieved the optimal comprehensive performance, demonstrating that each term in the loss function contributes effectively to the model's convergence and accuracy.

\begin{table}[t]
	\caption{Ablation study on adaptive loss function(Eq.~\ref{adaptive-loss})}
	\centering
	\setlength{\tabcolsep}{1.5mm}{
		\begin{tabular}{lccc}
			\toprule[1pt]
			\multirow{2}{*}{Method} & CREMA-D & AVE & KineticSound \\ 
			& Acc (\%) & Acc (\%) & Acc (\%)\\
			\hline
			\textbf{Ours (Full Model)} & \textbf{80.65} & \textbf{70.40} & \textbf{72.61}\\
			\hline
			Ours w/o $\mathcal{L}_{f}$ ($\alpha=0$) & 76.88 & 63.93 & 71.24\\
			Ours w/o $|g_i|^2$ ($\beta=0$) & 79.30 & 68.66 & 71.55 \\
			Ours w/o $|g_i - u_0|^2$ ($\gamma=0$) & 78.63 & 68.66 & 72.42\\
			Ours w/o $w_{i,0}, w_{i,1}$  &  77.42  &  67.16 & 71.12\\ 
			Ours w/o $\mathcal{L}_a,\mathcal{L}_v$   &  79.44 & 68.91 & 70.10 \\
			\bottomrule[1pt]
		\end{tabular}
	} % 对应 \setlength 的 }
\vspace{-0.5em}
\label{loss-ablation}
\end{table}

\subsection{Supplementary experiment}
\subsubsection{Dynamic evolution of modality gap}
\begin{figure*}[ht]
	\centering
	\begin{subfigure}[b]{0.25\textwidth}
		\centering
		\includegraphics[width=\textwidth]{figures/cremad-modalitygap-0.pdf}
		\caption{CREMAD: adaptive training stage 0}
		\label{fig:cremad-stage0}
	\end{subfigure}
	\hspace{1.2em}
	\begin{subfigure}[b]{0.25\textwidth}
		\centering
		\includegraphics[width=\textwidth]{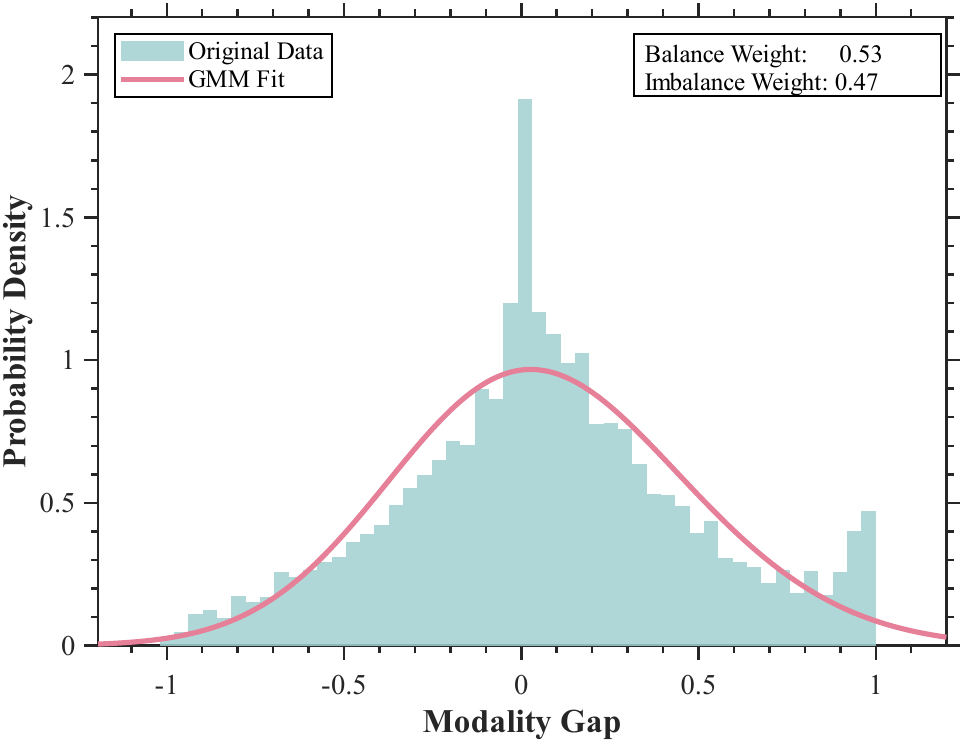}
		\caption{CREMAD: adaptive training stage 1}
		\label{fig:cremad-stage1}
	\end{subfigure}
	\hspace{1.2em}
	\begin{subfigure}[b]{0.25\textwidth}
		\centering
		\includegraphics[width=\textwidth]{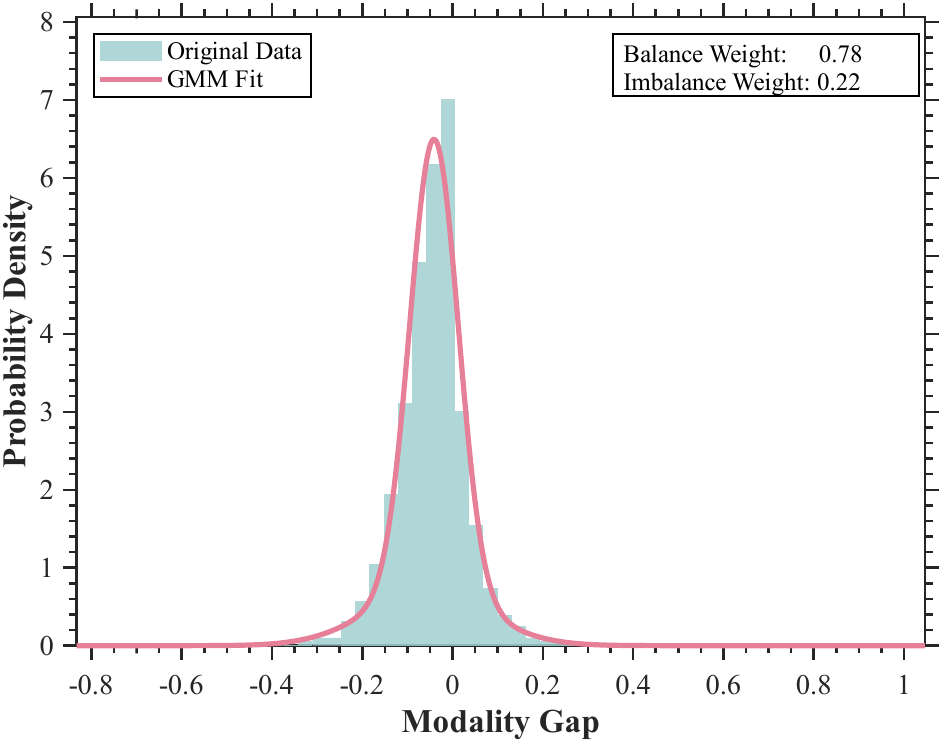}
		\caption{CREMAD: adaptive training stage 2}
		\label{fig:cremad-stage2}
	\end{subfigure}

    %--------------------------------------------------
    % Row 2: KineticSound dataset
    %--------------------------------------------------
	\begin{subfigure}[b]{0.25\textwidth}
		\centering
		\includegraphics[width=\textwidth]{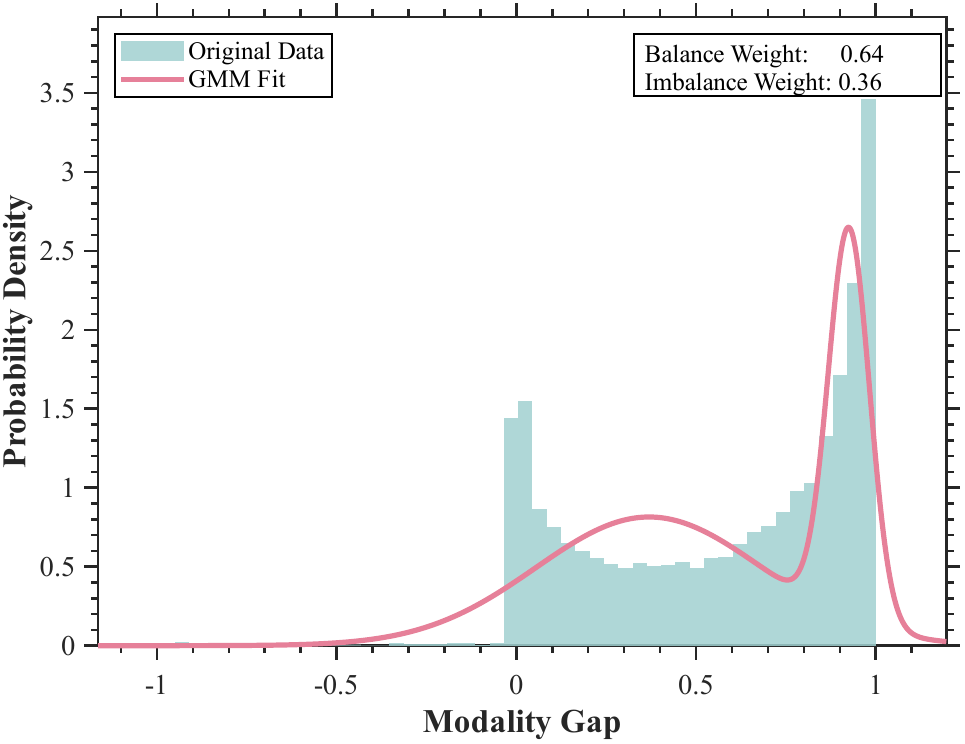}
		\caption{KS: adaptive training stage 0}
		\label{fig:ks-stage0}
	\end{subfigure}
	\hspace{1.2em}
	\begin{subfigure}[b]{0.25\textwidth}
		\centering
		\includegraphics[width=\textwidth]{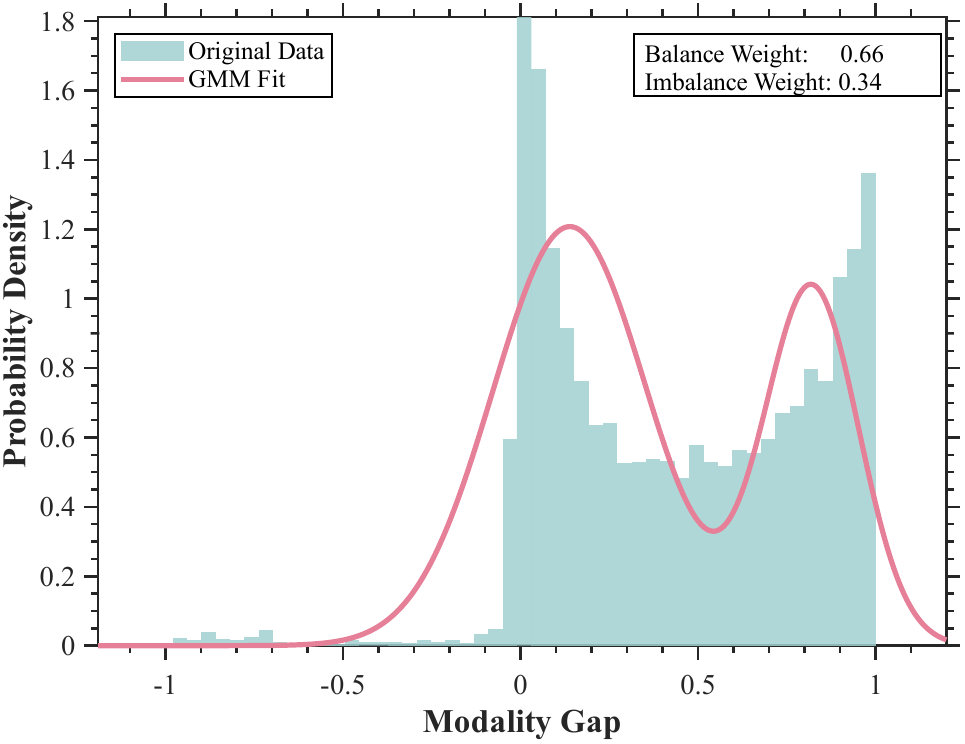}
		\caption{KS: adaptive training stage 1}
		\label{fig:ks-stage1}
	\end{subfigure}
	\hspace{1.2em}
	\begin{subfigure}[b]{0.25\textwidth}
		\centering
		\includegraphics[width=\textwidth]{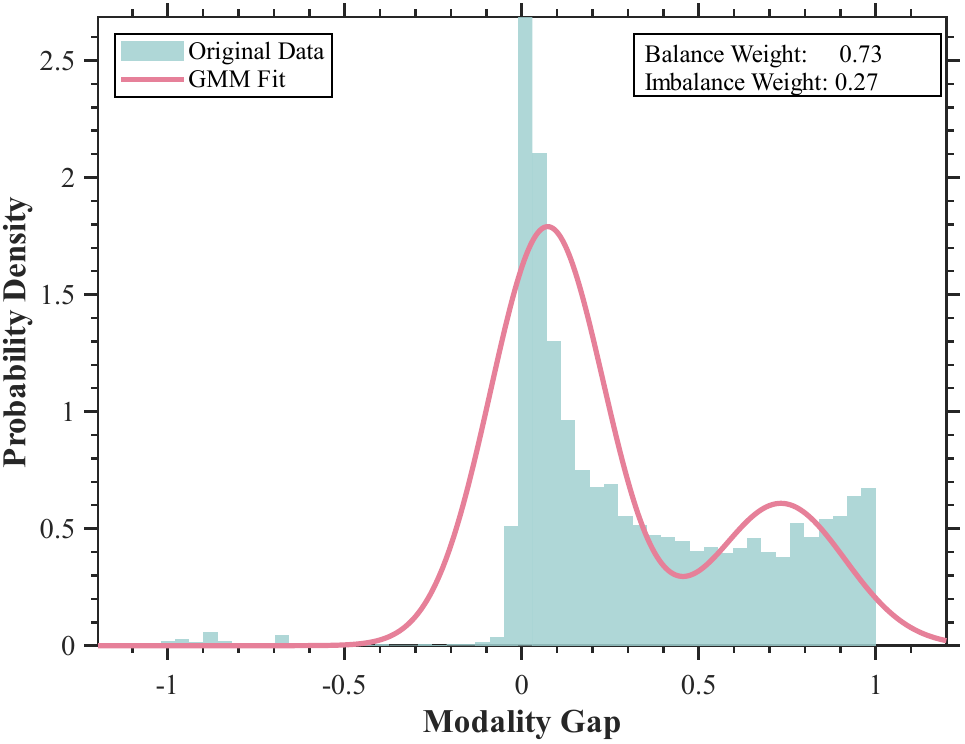}
		\caption{KS: adaptive training stage 2}
		\label{fig:ks-stage2}
	\end{subfigure}

    %--------------------------------------------------
    % Row 3: AVE dataset
    %--------------------------------------------------
	\begin{subfigure}[b]{0.25\textwidth}
		\centering
		\includegraphics[width=\textwidth]{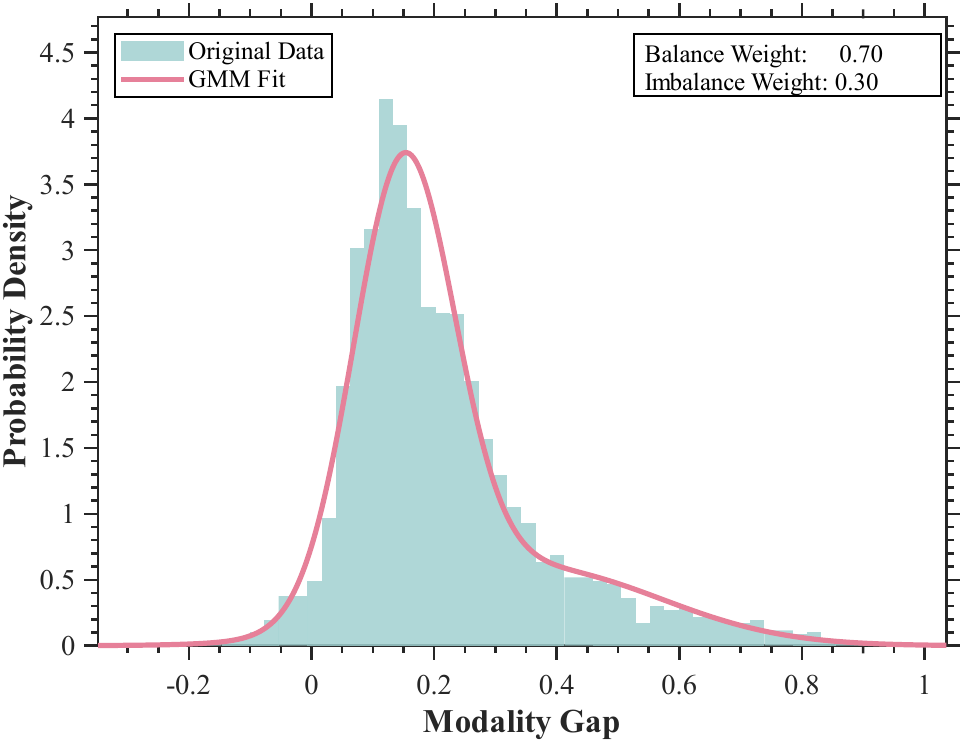}
		\caption{AVE: adaptive training stage 0}
		\label{fig:ave-stage0}
	\end{subfigure}
	\hspace{1.2em}
	\begin{subfigure}[b]{0.25\textwidth}
		\centering
		\includegraphics[width=\textwidth]{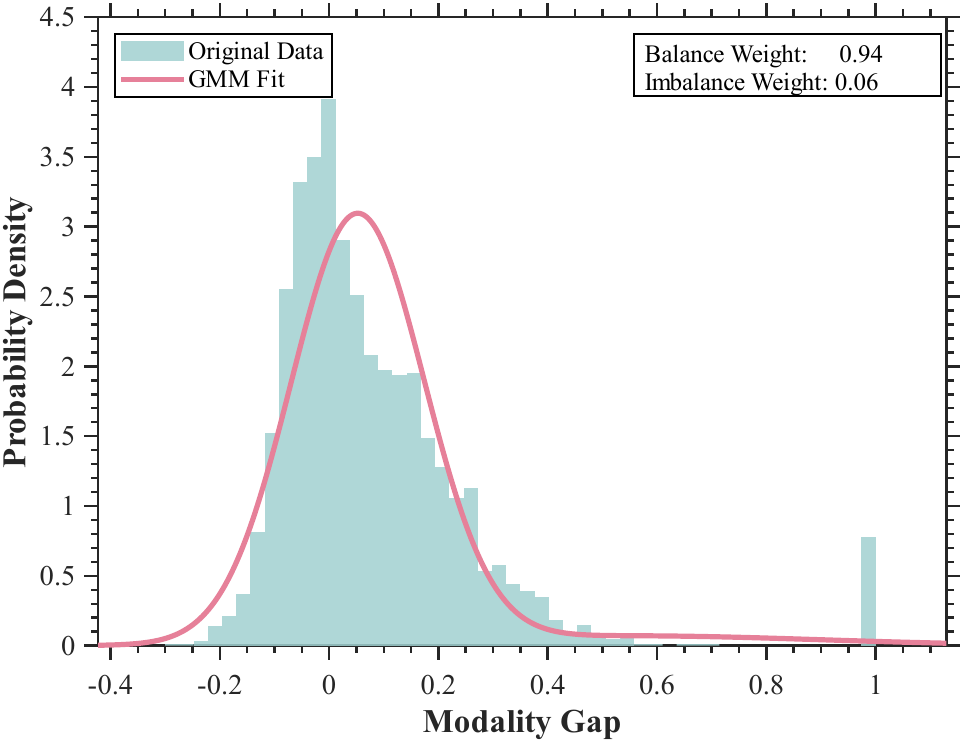}
		\caption{AVE: adaptive training stage 1}
		\label{fig:ave-stage1}
	\end{subfigure}
	\hspace{1.2em}
	\begin{subfigure}[b]{0.25\textwidth}
		\centering
		\includegraphics[width=\textwidth]{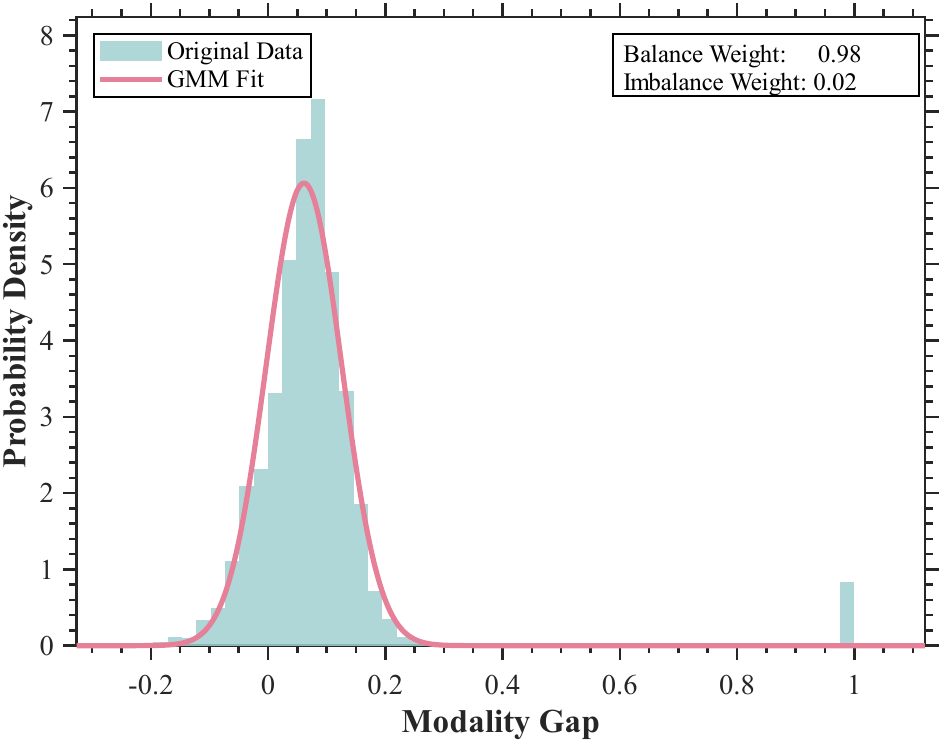}
		\caption{AVE: adaptive training stage 2}
		\label{fig:ave-stage2}
	\end{subfigure}

    \caption{Evolution of modality gap distributions across different datasets during adaptive training}
	\label{fig:GM-all-datasets}
\end{figure*}

GMM fitting and adaptive training are conducted in an alternating manner. We track the dynamic evolution of the modality gap distribution throughout training, as illustrated in Fig.~\ref{fig:GM-all-datasets}. For the CREMAD dataset, as training progresses, the two subgroups in the modality gap distribution exhibit a significant merging trend. The imbalanced subgroup gradually diminishes, suggesting that a substantial portion of imbalanced samples are effectively guided back toward the balanced manifold. Finally, the overall distribution converges to a sharply concentrated unimodal peak around origin, indicating a significant enhancement in cross-modal consistency. In contrast, for the KineticSound dataset, the balanced subgroup shifts noticeably away from zero after the warm-up stage, and the modality gap distribution retains a bimodal structure even in the later training phases. This suggests that the audio modality is considerably stronger than the visual modality, resulting in a large proportion of samples exhibiting imbalance that hinders effective fusion. Nevertheless, during the adaptive stage, the overall distribution still shifts leftward, and the proportion of balanced samples increases.

\subsubsection{Unimodal accuracy}

\begin{figure}[ht]
	\centering
	\begin{subfigure}[b]{0.23\textwidth}
		\centering % 子图内部居中
		\includegraphics[width=\textwidth]{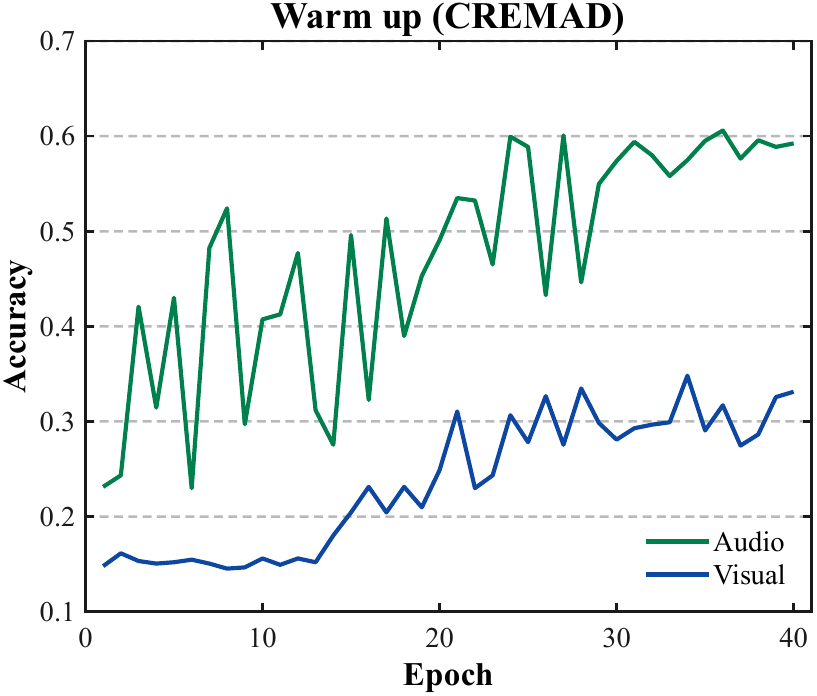}
		\caption{Warm up(CREMA-D)}
		\label{CREMAD-unimodal-acc:warmup}
	\end{subfigure}
	\hfill
	\begin{subfigure}[b]{0.23\textwidth} 
		\centering 
		\includegraphics[width=\textwidth]{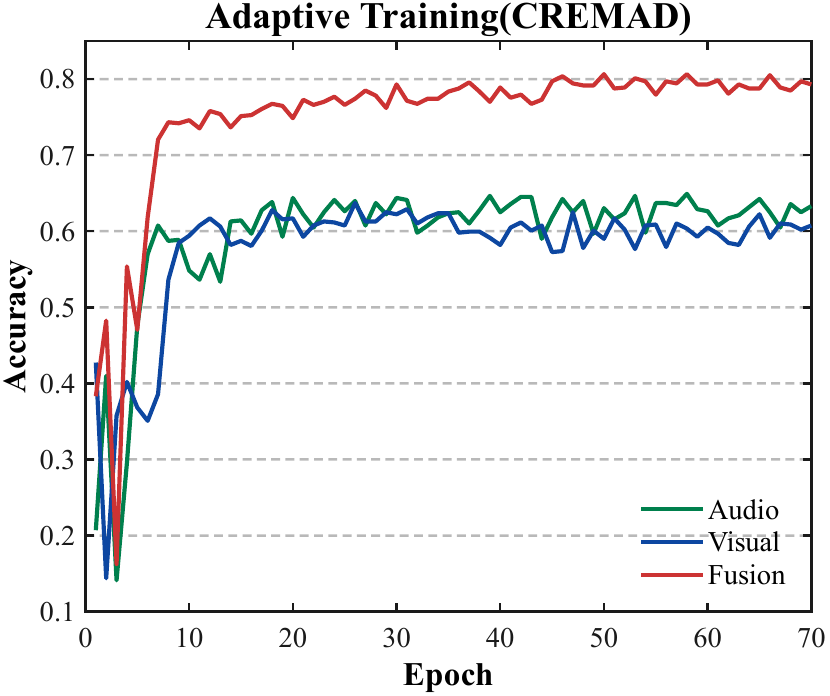}
		\caption{Adaptive training(CREMA-D)}
		\label{CREMAD-unimodal-acc:adaptive} 
	\end{subfigure}

	\begin{subfigure}[b]{0.23\textwidth}
		\centering 
		\includegraphics[width=\textwidth]{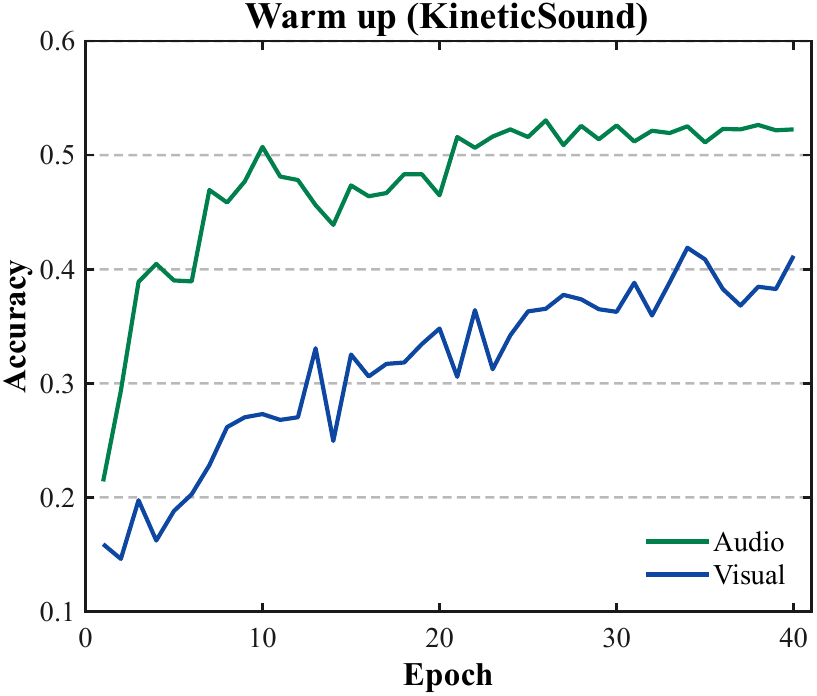} 
		\caption{Warm up(KineticSound)}
		\label{KS-unimodal-acc:warmup} 
	\end{subfigure}
	\hfill
	\begin{subfigure}[b]{0.23\textwidth}
		\centering 
		\includegraphics[width=\textwidth]{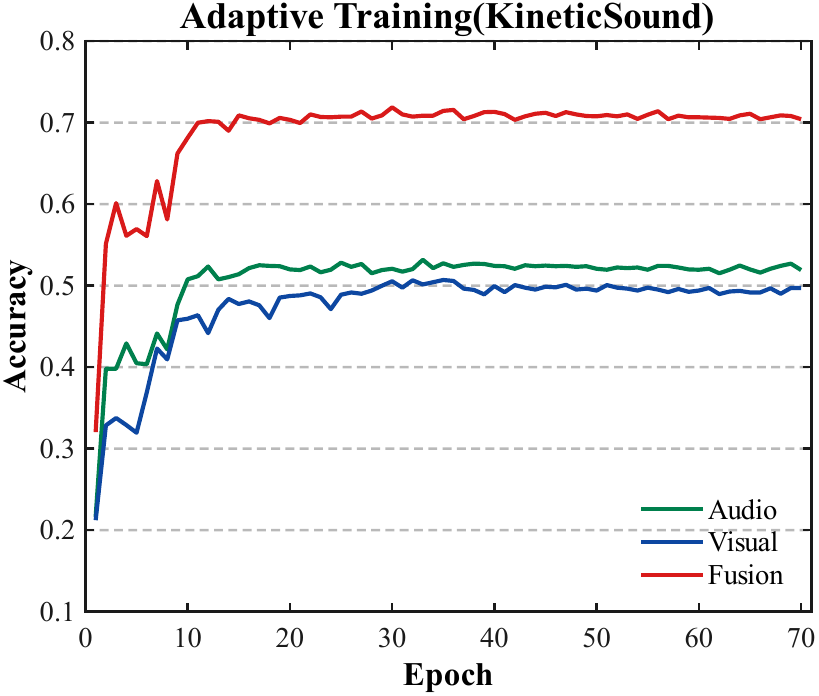}
		\caption{Adaptive training(KineticSound)}
		\label{KS-unimodal-acc:adaptive} 
	\end{subfigure}

	\begin{subfigure}[b]{0.23\textwidth}
		\centering 
		\includegraphics[width=\textwidth]{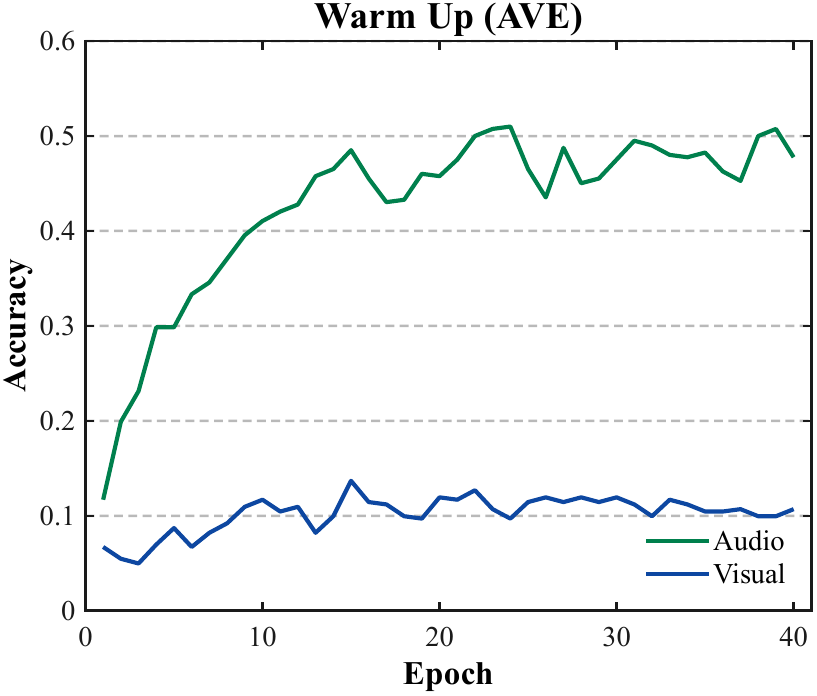} 
		\caption{Warm up(AVE)}
		\label{AVE-unimodal-acc:warmup} 
	\end{subfigure}
	\hfill
	\begin{subfigure}[b]{0.23\textwidth}
		\centering 
		\includegraphics[width=\textwidth]{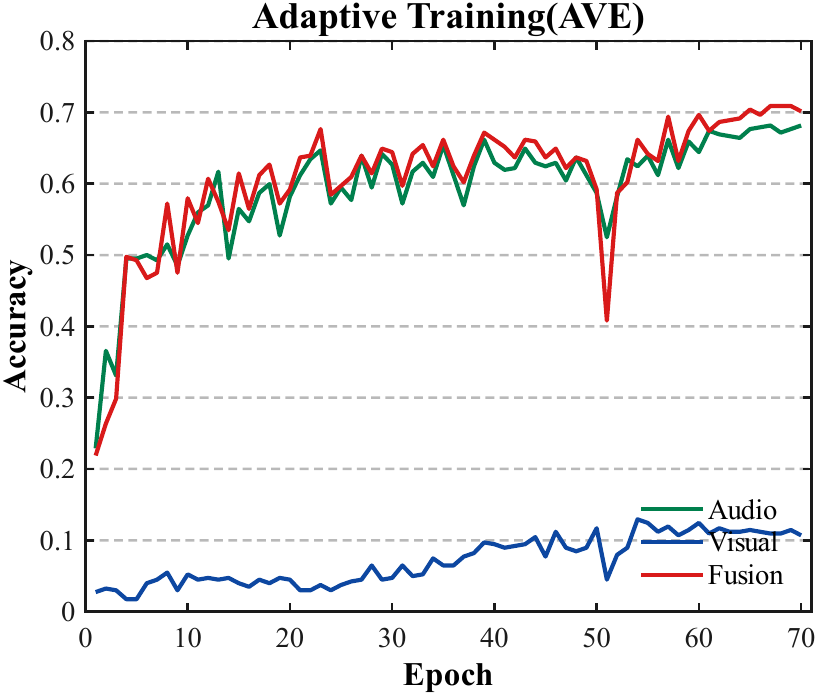}
		\caption{Adaptive training(AVE)}
		\label{AVE-unimodal-acc:adaptive} 
	\end{subfigure}

	\vspace{-0.5em} % 你原有的垂直间距

	\caption{Adaptive Training Improves Unimodal and Multimodal Accuracy}
	\label{warmup-adaptive-unimdal-acc} 
	
	\vspace{-0.5em}
\end{figure}

We further documented the dynamic evolution of unimodal accuracy curves during training, as illustrated in Fig.~\ref{warmup-adaptive-unimdal-acc}. During the warm-up phase, due to the absence of effective balancing constraints, the model exhibits severe modality dominance, resulting in significant prediction discrepancies between modalities (ranging from 10\% to 20\%). Upon entering the Adaptive Training phase, facilitated by the modality gap rectification mechanism, this state of imbalance is rapidly mitigated. The accuracy of the weaker modality shows a marked increase. This alignment further drives an improvement in fusion prediction accuracy, demonstrating that adaptive training not only bridges the performance gap between modalities but also further optimizes multimodal fusion features.

\subsubsection{Optimizer}
We further investigated the impact of optimizer strategies on the model's final performance, as presented in Table~\ref{tab:optimizer}. First, experiments indicate that resetting the optimizer state when transitioning from the warm-up to the adaptive training phase is critical. We attribute this to the fundamental disparity in optimization objectives between the two phases. Inheriting the warm-up optimizer state (denoted as "Same State" in Table~\ref{tab:optimizer}) carries over past momentum. This outdated momentum then disrupts gradient updates in the second phase. Second, regarding the choice of optimizer, we compared SGD with Adam~\cite{kingma2014adam} and observed that SGD yielded superior performance.
\begin{table}[t]
	\centering
	\caption{Ablation study on optimizer settings and states.}
	\begin{tabular*}{\columnwidth}{l@{\extracolsep{\fill}}lcc}
		\toprule[1pt]

		\multirow{2}{*}{Optimizer Setting} & CREMA-D & AVE \\
		& Accuracy (\%) & Accuracy (\%) \\
		\hline
		SGD (Reset State) & 80.65 & 70.40 \\
		Adam (Reset State) & 79.91 & 70.19 \\
		\hline 
		SGD (Same State) & 78.76 & 67.91 \\
		Adam (Same State) & 75.67 & 66.67 \\
		\bottomrule[1pt]
	\end{tabular*}

	\vspace{-0.5em}
	\label{tab:optimizer}
\end{table}

\subsubsection{Another Distributions}
To comprehensively evaluate the sensitivity of model performance to prior distribution assumptions, we broadened our approach to modeling the modality gap distribution. In addition to the Gaussian Mixture Model (GMM), we introduced candidate distributions with distinct shape characteristics for comparison. The Student's $t$-Mixture Model (SMM) is designed to enhance the fitting capability for heavy-tailed imbalanced samples, while the Laplace Mixture Model (LMM) is employed to better capture the dense balanced samples concentrated at the origin. Specifically, given the set of modality gaps $\mathcal{G} = \{ \hat{g}_i \}_{i=1}^N$, we fit each candidate distribution while maintaining the adaptive optimization workflow consistent with the established framework. The detailed comparative results are summarized in Table~\ref{tab:distribution_ablation}. We observe that the Laplace Mixture Model (LMM) generally yields suboptimal performance across all three datasets. This outcome may be attributed to the sharp discontinuity and lack of smooth transitions in the Laplace probability density, which undermines the stability required for adaptive optimization.

\begin{table}[t]
	\caption{Performance comparison under different distribution fitting assumptions on the \textbf{CREMA-D}, \textbf{AVE}, and \textbf{KineticSound} datasets. GM, SM, and LM denote Gaussian Mixture, Student's $t$-Mixture, and Laplace Mixture models, respectively.}
	
	\centering
	\begin{tabular*}{\columnwidth}{@{\extracolsep{\fill}}lccc}
		\toprule[1pt]
		\multirow{2}{*}{Distributions} 
		& \multicolumn{1}{c}{CREMA-D} 
		& \multicolumn{1}{c}{AVE} 
		& \multicolumn{1}{c}{KS} \\
		\cmidrule(lr){2-2} \cmidrule(lr){3-3} \cmidrule(lr){4-4}
		& Acc(\%) & Acc (\%) & Acc (\%) \\
		\hline
		GMM(Prob) & \textbf{80.65} & \textbf{70.40} & \textbf{72.61} \\
		GMM(KL) & 78.63 & 69.15 & 71.83 \\
		SMM(Prob) & 79.91 & 68.91 & 72.42 \\
		SMM(KL) & 78.63 & 68.91 & 72.26 \\
		LMM(Prob) & 77.15 & 70.15 & 72.14 \\
		LMM(KL) & 76.88 & 68.41 & 70.84 \\
		\hline
	\end{tabular*}
	
	\vspace{-0.5em}
	\label{tab:distribution_ablation}
\end{table}

%超参数敏感实验
\subsubsection{Hyperparameter Sensitivity Analysis}

In Eq.~\ref{adaptive-loss}, we introduce a dynamic annealing coefficient $\lambda_t = \rho^{\text{t}}$, designed to balance the modality gap constraint during early training with joint feature learning in later stages. To determine the optimal decay rate $\rho$, we performed a hyperparameter sensitivity analysis across the range $\rho \in \{1.00, 0.95, 0.90, 0.85, 0.80\}$, the results of which are summarized in Table~\ref{tab:sensitivity_lambda}. We observe that maintaining a constant strong constraint ($\rho=1.0$) leads to a significant drop in accuracy on CREMA-D to 76.34\%. This indicates that persistently imposing an excessive penalty on the modality gap interferes with the learning of fusion features, thereby compromising classification accuracy. Conversely, performance also declines on the AVE and CREMA-D datasets when the decay rate is too small. This is likely attributable to $\lambda_t$ decaying too rapidly to achieve effective modality gap rectification, failing to adequately mitigate the substantial noise induced by the modal prediction bias of imbalanced samples.

\begin{table}[t]
	\centering
	\caption{Effect of the annealing decay rate $\rho$. Classification accuracy (\%) on three datasets under different values of $\rho$ in the decay schedule $\lambda_t = \rho^t$.}
	\label{tab:sensitivity_lambda}
	\setlength{\tabcolsep}{3.5mm}
	\begin{tabular}{lccc}
		\toprule
		\textbf{Decay Rate ($\rho$)} & \textbf{CREMA-D} & \textbf{AVE} & \textbf{KS} \\
		\midrule
		1.00               & 76.34 & 68.66 & 72.46 \\
		0.95               & 80.65 & 70.40 & 72.42 \\
		0.90               & 79.03 & 69.65 & 72.61 \\
		0.85               & 78.63 & 67.41 & 72.06 \\
		0.80               & 78.36 & 67.41 & 72.61 \\
		\bottomrule
	\end{tabular}
\end{table}

\subsubsection{Data Purification and FT}

We further explore the potential of GMM-based probabilistic modeling for data selection. Our findings indicate that the high-quality balanced subset identified by the GMM can function as an effective data filtering strategy. Even without introducing the adaptive loss, fine-tuning the warm-up model exclusively on these selected subsets yields substantial performance gains. Specifically, based on Eq.~\ref{GM-fitting-w0}, we adopt the posterior probability $w_{i,0}$ as the selection criterion, which indicates how likely a sample is to belong to the balanced subgroup. By applying different confidence thresholds $\tau \in \{50\%, 60\%, \dots, 95\%\}$, we construct high-quality data subsets and continue training the model using only the warm-up loss defined in Eq.~\ref{warm-up-loss}. As reported in Table~\ref{tab:finetuning}, under all threshold settings, the fine-tuned models consistently outperform the warm-up baseline trained on the full dataset. The experimental results reveal a critical trade-off mechanism. Although increasing the threshold enhances modality balance within the selected subset, it inevitably reduces data diversity. Taking KineticSound as an example, the best performance is achieved at $\tau = 60\%$, suggesting that overly aggressive data filtering is suboptimal. In practice, the generalization capability of deep learning models often benefits from the implicit regularization effect introduced by moderately imbalanced data.

\begin{table}[t]
	\centering
	\caption{Performance comparison (Acc. \%) on high-quality subsets selected via confidence thresholds ($w_{i,0}$). The Data column denotes the percentage of samples retained from the original dataset.}

	\setlength{\tabcolsep}{4pt} 
	
	\sisetup{table-format=3.2}
	
	% --- 注意：去掉了 \resizebox{...}{!}{...} 包裹 ---
	\begin{tabular}{l
			S[table-format=3.2]
			S[table-format=2.2]
			S[table-format=3.2]
			S[table-format=2.2]
			S[table-format=3.2]
			S[table-format=2.2]
		}
		\toprule[1.5pt]
		
		\multirow{2}{*}{$w_{i,0}$} & \multicolumn{2}{c}{CREMA-D} & \multicolumn{2}{c}{AVE} & \multicolumn{2}{c}{KS} \\
		
		\cmidrule(lr){2-3} \cmidrule(lr){4-5} \cmidrule(lr){6-7}
		
		& {Data} & {Acc.} & {Data} & {Acc.} & {Data} & {Acc.} \\
		
		\midrule
		
		Baseline & 100.00 & 67.47 & 100.00 & 64.48 & 100.00 & 65.54 \\
		
		\addlinespace[0.4em] % 稍微增加一点行间距，提升质感
		
		$\ge 50\%$ & 86.04 & \bfseries 77.69 & 75.21 & 66.42 & 51.93 & 66.92 \\
		$\ge 60\%$ & 85.35 & 75.27 & 74.00 & 66.92 & 51.06 & \bfseries 67.62 \\
		$\ge 70\%$ & 84.61 & 75.13 & 72.37 & 66.42 & 50.32 & 67.19 \\
		$\ge 80\%$ & 83.71 & 75.00 & 69.87 & \bfseries 67.16 & 49.16 & 66.72 \\
		$\ge 90\%$ & 82.35 & 75.94 & 63.71 & \bfseries 67.16 & 47.90 & 66.92 \\
		$\ge 95\%$ & 80.89 & 75.81 & 49.55 & 67.16 & 46.88 & 66.56 \\
		\bottomrule[1.5pt]
	\end{tabular}
	\label{tab:finetuning}
\end{table}

\section{Conclusion}
To address the challenge of multimodal imbalance in multimodal learning, we propose a sample-level adaptive optimization method integrated into the second stage of a two-stage training framework. Specifically, we define the modality gap as a metric to quantify sample-level multimodal imbalance, formulated based on Softmax probability differences and KL divergence. By modeling the distribution of modality gap, we reveal the existence of distinct balanced and imbalanced subgroups within the multimodal data. We employ a Gaussian Mixture Model (GMM) to fit the modality gap distribution, utilizing this as a prior, we compute the posterior probabilities of individual samples belonging to either the balanced or imbalanced subgroup via Bayes' theorem. These posterior probabilities are subsequently incorporated into an adaptive loss function, enabling differentiated optimization for balanced and imbalanced samples. Experimental results demonstrate that our method effectively resolves the multimodal imbalance problem and exhibits robust generalization capabilities.

\bibliographystyle{IEEEtran}
\bibliography{references}
\end{document}